\documentclass{article}

\usepackage[final]{neurips_2025}

\usepackage[breakable]{tcolorbox}
\definecolor{mBlue}{HTML}{173b4f}
\usepackage{listings}
\usepackage{float}

\usepackage[utf8]{inputenc} 
\usepackage[T1]{fontenc}    
\usepackage{hyperref}       
\usepackage{url}            
\usepackage{booktabs}       
\usepackage{amsfonts}       
\usepackage{nicefrac}       
\usepackage{microtype}      
\usepackage{xcolor}         
\usepackage{amsmath}
\usepackage{amsthm}
\usepackage{amssymb}
\usepackage{mathtools}
\usepackage{mdframed}
\usepackage{tikz}
\usepackage{caption}
\usepackage{subfigure}

\usepackage{graphicx}
\usepackage{textcomp}

\definecolor{schemaBlue}{HTML}{c9e0ff}
\definecolor{babyblue}{rgb}{0.54, 0.81, 0.94}
\definecolor{blue(pigment)}{rgb}{0.2, 0.2, 0.6}

\newtcolorbox{custombox}{
  colframe=pink!80!red, 
  colback=pink!20, 
  arc=5pt, 
  boxrule=1.5pt, 
  coltitle=black, 
  colbacktitle=pink!80!red, 
  fonttitle=\bfseries,
  title={\textbf{Model identification prompt}},
  breakable 
}

\newtcolorbox{custombox_decision_making}{
  colframe=pink!80!red, 
  colback=pink!20, 
  arc=5pt, 
  boxrule=1.5pt, 
  coltitle=black, 
  colbacktitle=pink!80!red, 
  fonttitle=\bfseries,
  title={\textbf{Decision making prompt}},
  breakable 
}

\newtcolorbox{custombox_learning}{
  colframe=pink!80!red, 
  colback=pink!20, 
  arc=5pt, 
  boxrule=1.5pt, 
  coltitle=black, 
  colbacktitle=pink!80!red, 
  fonttitle=\bfseries,
  title={\textbf{Learning prompt}},
  breakable 
}

\newtcolorbox{custombox_planning}{
  colframe=pink!80!red, 
  colback=pink!20, 
  arc=5pt, 
  boxrule=1.5pt, 
  coltitle=black, 
  colbacktitle=pink!80!red, 
  fonttitle=\bfseries,
  title={\textbf{Planning prompt}},
  breakable 
}

\newtcolorbox{custombox_wm}{
  colframe=pink!80!red, 
  colback=pink!20, 
  arc=5pt, 
  boxrule=1.5pt, 
  coltitle=black, 
  colbacktitle=pink!80!red, 
  fonttitle=\bfseries,
  title={\textbf{Working memory prompt}},
  breakable 
}

\title{Generating Computational Cognitive Models \\ using Large Language Models}

\author{
  Milena Rmus $\dagger$\\
  Helmholtz Munich \\
  \texttt{milena.rmus@helmholtz-munich.de} \\
  \And
  Akshay K. Jagadish $\dagger$ \\
  Princeton University \\
  \texttt{akshay.jagadish@princeton.edu} \\
  \AND
  Marvin Mathony \\
  Helmholtz Munich \\
  \texttt{marvin.mathony@helmholtz-munich.de} \\
  \And
  Tobias Ludwig \\
  Tübingen University \\
  \texttt{tobias.ludwig@uni-tuebingen.de} \\
  \And
  Eric Schulz \\
  Helmholtz Munich \\
  \texttt{eric.schulz@helmholtz-munich.de} \\
  \\
  $\dagger$ Milena Rmus and Akshay K. Jagadish contributed equally.
}

\begin{document}

\maketitle

\begin{abstract}
Computational cognitive models, which formalize theories of cognition, enable researchers to quantify cognitive processes and arbitrate between competing theories by fitting models to behavioral data. Traditionally, these models are handcrafted, which requires significant domain knowledge, coding expertise, and time investment. However, recent advances in machine learning offer solutions to these challenges. In particular, Large Language Models (LLMs) have demonstrated remarkable capabilities for in-context pattern recognition, leveraging knowledge from diverse domains to solve complex problems, and generating executable code that can be used to facilitate the generation of cognitive models. 
Building on this potential, we introduce a pipeline for Guided generation of Computational Cognitive Models (GeCCo). Given task instructions, participant data, and a template function, GeCCo prompts an LLM to propose candidate models, fits proposals to held-out data, and iteratively refines them based on feedback constructed from their predictive performance. We benchmark this approach across four different cognitive domains -- decision making, learning, planning, and memory -- using three open-source LLMs, spanning different model sizes, capacities, and families. On four human behavioral data sets, the LLM generated models  consistently matched or outperformed the best domain-specific models from the cognitive science literature. 
To validate these findings, we performed control experiments that investigated (1) the contribution of the different LLM features (model size, model family, capacities); (2) the causal role of different prompt components; (3) the effect of data contamination; (4) the ability to recover ground truth models from simulated data; and (5) the total explainable variance in human behavior captured by LLM-generated models. 
Taken together, our results suggest that LLMs can rapidly generate cognitive models with conceptually plausible models that rival -- or even surpass -- the best models from the literature across diverse task domains. The code for GeCCo is available at \url{https://github.com/MilenaCCNlab/gecco.git}
\end{abstract}

\section{Introduction}

Across the sciences, computational models provide a formal framework for understanding natural phenomena. In cognitive science, these models ground theories about the cognitive processes that underlie human behavior. In practice, researchers hand-craft these models, fit them to behavioral data, and iteratively refine them until they capture behavior sufficiently well \citep{polk2002cognitive}. 
However, handcrafting cognitive models that accurately explain behavior can be time-consuming \citep{musslick2024closed, musslick2024automatingpracticescience}. It requires trained researchers to conduct lengthy literature reviews, formulate cognitively plausible theories, and implement computationally feasible algorithms to evaluate these theories. While assumptions are necessary to constrain any model, researchers' own theoretical commitments and expertise may unintentionally narrow the range of models they consider \citep{krefeld2022structural, taatgen2016cognitive}, potentially missing better explanations of the data \citep{frischkorn2018cognitive, addyman2012computational}. 


Recent progress in Large Language Models (LLMs) offers a practical route to addressing these challenges, tackling both efficiency and biases, while broadening the space of possible models \citep{wang2023scientific, binz2023should, musslick2024automatingpracticescience, li2024automated}. Specifically, LLMs exhibit at least three abilities that can facilitate the development of a flexible, domain-general framework for generating cognitive models. First, LLMs can process behavioral data formatted in natural language along with the corresponding task description \citep{schubert2024context, jagadish2024human}, providing them with the flexibility to handle diverse task domains with varying levels of complexity \citep{binz2024centaur}. Second, they can identify patterns in complex problems using domain knowledge in-context, and generate hypotheses about the data-generating process \citep{xiao2024verbalized}. 
Third, their ability to synthesize highly accurate programs \citep{austin2021program,perez2021automatic} lends itself nicely to cognitive model generation and subsequent evaluation. Importantly, these three abilities have already shown promise in statistical modeling, where LLMs have been used to generate and evaluate models represented as probabilistic programs \citep{li2024automated}.


\vspace{+0.15cm}

\begin{figure*}[h]
    \centering
    \includegraphics[width=\textwidth, keepaspectratio]{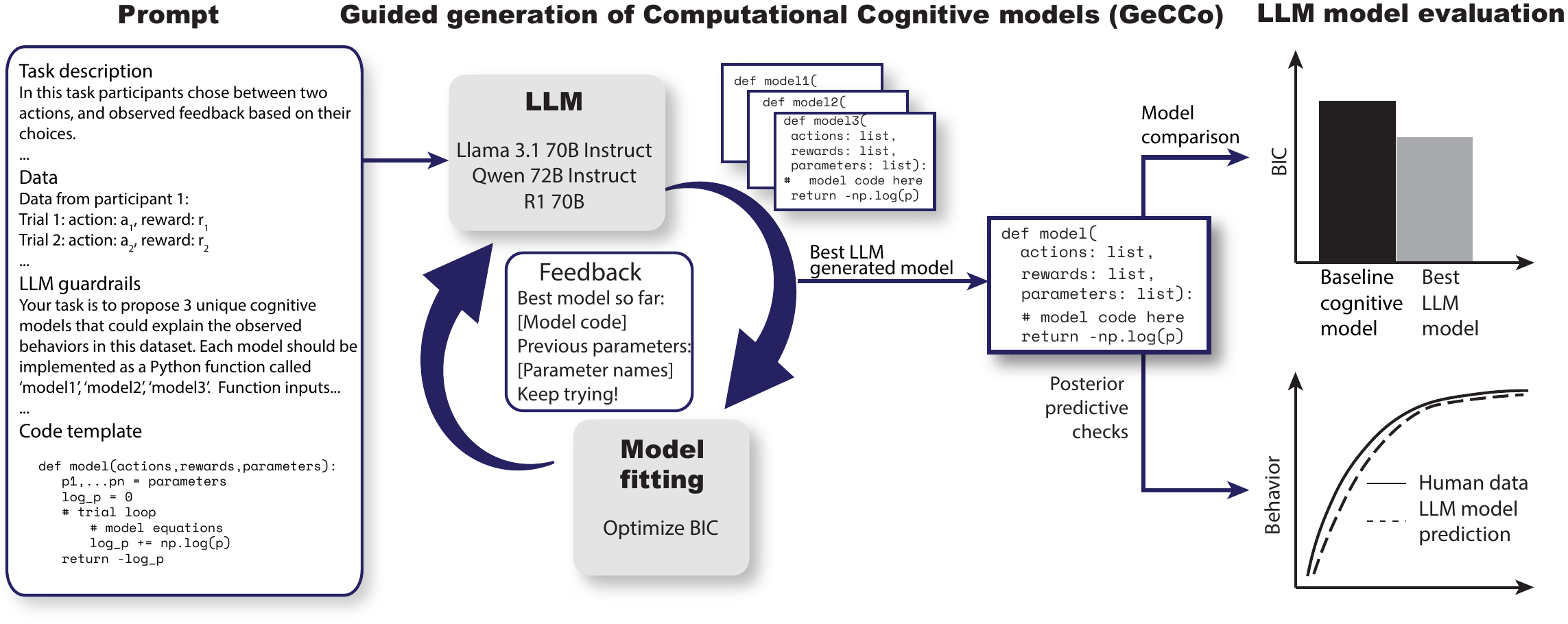}
    \caption{Schematic of GeCCo: We prompt the LLM with a task description, participant data, guardrails to constrain the format of LLM responses, and the code template to generate cognitive models that offer different explanations of the underlying data as Python functions. Model generation evolves over 10 sampling iterations. During each iteration, three LLM-generated models are fitted offline to the held out data (not included in the prompt), and the fitness metric Bayesian Information Criterion (BIC; \citealt{watanabe2013widely}) is used to provide feedback to the LLM on the subsequent iteration. The best model across all 10 sampling iterations is used for evaluation. The LLM-generated models are evaluated by 1) fitting them to behavioral data and comparing the model fit to that of the baseline cognitive model (e.g. the best performing model from the literature) using  BIC, and 2) running posterior predictive checks - i.e. simulating the models and comparing simulated to ground truth data - to further verify their validity. For full prompts, see the Appendix  \ref{all_prompts}.
    }
    \label{llm-prompt}
\end{figure*}

In this work, we leveraged these LLM abilities to generate hypotheses about the cognitive processes underlying human behavior. We developed a novel pipeline for Guided generation of Computational Cognitive Models (GeCCo) that helps synthesize cognitive models as Python functions using LLMs, and iteratively refines them based on feedback on their predictive performance. 
We evaluated this approach across four different cognitive domains: decision making, learning, planning, and memory, using three open-source LLMs, with different features including model sizes, capacities, and families.
Specifically, we first compared LLM-generated models with the best domain-specific model from the literature in terms of predictive performance and then validated the best performing LLM model using posterior predictive checks.
Furthermore, we conducted control experiments to gain a mechanistic understanding of this pipeline. These included: (1) linear mixed-effects models to isolate the contribution of different features of the base LLM; (2) LLM prompt ablations to test the causal role of instruction, data, iterative feedback, and template components; (3) LogProber \citep{yax2024assessing} estimation of data contamination in prompts; (4) simulations to access recovery of ground truth models; and (5) a comparison with a foundation model of cognitive science (\textsc{Centaur}; \citealt{binz2024centaur}) to approximate the total explainable variance captured in human behavior. Across all domains, we found that LLM‑generated models matched or surpassed the best domain-specific model from the literature, and captured as much explainable variance as \textsc{Centaur}, while remaining interpretable. The results from the control experiments revealed that the iterative feedback component in the prompt and the reasoning ability of the LLMs are the two main drivers of this performance, with no traces of contamination in the prompt. Together, these results position GeCCo as a practical, scalable approach for rapid cognitive‑model generation.


\section{Methods}


\paragraph{Guided generation of Computational Cognitive Models (GeCCo).}

We implemented GeCCo, a structured guided sampling pipeline for generating cognitive models with LLMs. Each pipeline run consisted of 10 sampling iterations, repeated across 5 independent runs per domain to assess stability. On each iteration, the base LLM was prompted using a fixed prompt structure (see Appendix \ref{llm-prompt-structure}) that included 1) a natural language task description, 2) behavioral data from a subset of participants (prompt data), 3) instructions and guardrails (e.g., write a Python function with a specific name and defined inputs/outputs), 4) a domain-specific template model used primarily for syntactic guidance, and 5) feedback (included from iteration 2 onward). Using this prompt, the LLM generated three distinct cognitive models with non-overlapping parameter sets. These models were parsed from the LLM response, converted into executable Python functions (e.g., via exec(function\_string)), and their parameter names were extracted. Each model was then fit to a second, held-out validation dataset (not shown in the prompt) using the minimize function from the SciPy optimization library \citep{2020SciPy-NMeth}, initialized from 20 random starting points to avoid local minima. Model fit was evaluated using the Bayesian Information Criterion (BIC), which was used to identify the current best-performing model. This best model, and a list of previously used parameter names were then included in the next prompt to guide model generation and avoid duplicates. If model generation failed (e.g., due to syntax or runtime errors), the iteration was automatically restarted. After completing all iterations in a run, the best LLM-generated model compared against competing cognitive models by evaluating on a third, held-out test set. See Figure \ref{llm-prompt} for an overview of the pipeline and Appendix \ref{all_prompts} for exact prompts that were used in each of the experiments.

\paragraph{LLM specifications.}
We used three open-source LLMs: Llama-3.1-Instruct-70B (Llama; \citealt{meta2024llama3.1}, DeepSeek-R1-Distill-Llama-3.1-70B (R1; \citealt{guo2025deepseek}), and Qwen2.5-72B-Instruct (Qwen; \citealt{yang2024qwen2}). We chose these models as they span different sizes (70B, 72B), capabilities (with/without reasoning), and families (Qwen, Llama, and Deepseek).  Importantly, all of our tests were performed in-context. We set the temperature to 0.2 for Llama models, 0.15 for Qwen and 0.1 for R1 to encourage some exploration when generating models. In practice, our approach takes a maximum of 8 hours per task domain on four Nvidia A100s with 40GB memory each. 

\paragraph{Evaluation.} 
We applied GeCCo to four canonical cognitive domain paradigms: decision making, learning, planning, and memory. These paradigms include well-established baseline models that are easy to fit and support systematic increases in complexity, both in parameter count and in the cognitive processes they formalize. We began with a simple decision-making task, progressed to sequential learning with trial dependencies, then examined a planning task involving multi-step reasoning, and finally used a task that examines interactions between working memory and reinforcement learning. Evaluating LLM-generated models on human behavioral datasets from these paradigms, where data are noisy and no single model fully captures the underlying processes, is essential, as it reflects the real-world settings in which cognitive scientists would use our approach.
\paragraph{Metrics.}
We evaluated the performance of the GeCCo generated model using the Bayesian Information Criterion (BIC; \citealt{watanabe2013widely}), which balances goodness of fit and model complexity by penalizing the number of parameters. Because BIC is a relative metric, when models achieve similar fit, it favors the less complex (fewer parameter) model. The pipeline is agnostic to the model quality measure; for example, in Appendix \ref{aic_section}, we report results using the Akaike Information Criterion (AIC) as a metric. In the last stage of the evaluation, we additionally computed the exceedance probability (EXP; \citealt{stephan2009bayesian}), which quantifies the posterior probability that one model explains the data better than its competitors. EXP expresses, in probabilistic terms, how confident we are that a given model provides the most prevalent explanation of the observed behavior.

\section{Experiment 1: Decision making}

\paragraph{Task.}
To understand the mechanisms underlying human decision making, researchers have commonly resorted to multi-feature decision making tasks \citep{bogacz2006physics}. In such tasks, participants typically choose between multiple options, each defined by a distinct set of features. 
We considered the paradigm from \cite{hilbig2014generalized}, where participants chose between two options that are defined by four feature values and their respective validities (Fig. \ref{fig:decision_making}A; see Appendix \ref{app:human_experiments} for more details).

\paragraph{Baseline cognitive model.}

The winning model in the \cite{hilbig2014generalized} study was a probabilistic weighted additive model (pWADD), which combines feature values and inferred feature weights (instead of feature validities) to predict human choices. Critically, the inferred weights can match or completely differ from the feature validity provided by the experimenter; see Appendix \ref{sec:decision_making_cog_model}.

\vspace{+0.15cm}
\begin{figure*}[h]
    \centering
    \includegraphics[width=\textwidth, keepaspectratio]{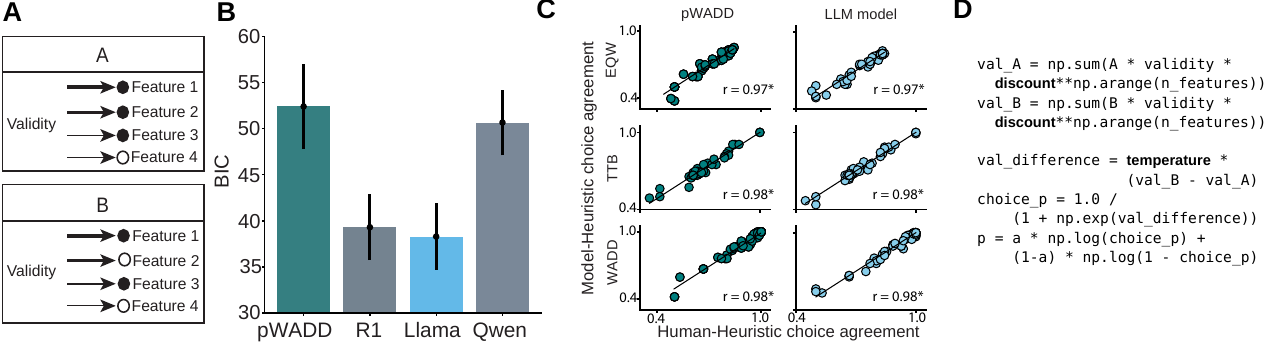}
    \caption{Experiment 1: Decision Making. A) Schematic of the decision task from \cite{hilbig2014generalized}, where participants were asked to choose between two options based on four binary features and their validities. Arrow thickness indicates validity value (i.e., thicker arrows mean higher validity). B) Model fit comparison: LLM-generated models from R1 and Llama outperformed the best literature model. C) Posterior predictive checks showed that proportions of choices accounted for by the canonical heuristics (Equal Weighting, Take The Best and Weighted Additive Heuristics) closely aligned between human data and respective LLM and pWADD model predictions. D) Code of the best LLM-generated model (Llama), which can arbitrate between three canonical heuristics mentioned above via a discount factor.}
    \label{fig:decision_making}
\end{figure*}

\paragraph{Results.}

We found that both R1 and Llama produced a model that outperformed the best literature model (Llama BIC$_{M \pm SEM}$ vs. pWADD BIC$_{M \pm SEM}$: 39.40 $\pm$ 4.54 vs. 51.97  $\pm$ 4.52; t-test (pWADD > Llama BIC: t(52) = 41.8, p < .001; see Figure \ref{fig:decision_making}B). The best LLM-generated model (by Llama) weighs the option features by the respective feature validities and a discount factor. The strength of this model is that it can arbitrate between Take the Best, Weighted Additive, and Equal Weighting decision making strategies with a single parameter; see Figure \ref{fig:decision_making}D. Although the pWADD model can do the same using its inferred feature weights, it requires four parameters instead of one parameter used by the LLM-generated model. To validate the LLM-generated model, we performed posterior predictive checks - which typically includes simulating data based on the cognitive model, and checking whether the simulated data agrees with the human data (for details, refer to Appendix \ref{posterior_predictive_checks}). We found that the data simulated from the best LLM-generated model aligned closely with human choices; see Figure \ref{fig:decision_making}C. 

\paragraph{Scientific Insight.}
The parsimonious way in which the LLM-generated model selects between heuristics aligns with the continuum-of-models view, which interprets heuristics as Bayesian inference under extreme priors \citep{parpart2018heuristics}. In fact, the discount term in the LLM-generated model parallels the prior-strength parameter in the constraint-optimal regression and half-ridge formulations from \cite{parpart2018heuristics}, both of which enable smooth modulation of posterior feature weights—and therefore the decision rule. Importantly, the LLM-generated model achieves this flexibility without adopting a Bayesian framework.

\section{Experiment 2: Learning}


\paragraph{Task.}
Multi-armed bandit tasks are frequently used in the study of feedback-driven learning \citep{frank2004carrot, pessiglione2006dopamine, wang2016learning}. Generally, in a bandit task, an agent chooses between \textit{N} arms, often with a predefined reward contingency associated with each arm. The agent receives feedback for their chosen action and, over a sequence of trials, learns to adjust action selection in a way that maximizes positive outcomes. For analysis, we considered the two-armed bandit task dataset from \cite{chambon2020information}. 
In addition to the classic condition (partial feedback), where only rewards for chosen options are observed (see Appendix \ref{partial_feedback} and Fig. \ref{fig:partial_feedback} for results), the authors include a full-feedback condition, where rewards for both unchosen and chosen options are observable; see Appendix \ref{app:human_experiments} for details. 

\paragraph{Baseline cognitive model.}

The winning model from \cite{chambon2020information} is a variant of the classic Rescorla-Wagner model that is commonly used to study learning dynamics in bandit tasks. This model ($RW^{4\alpha}$) contains four learning rates: learning rates for positive and negative prediction errors for chosen actions, and the same for unchosen actions. 
In addition to using a delta rule for action-value updating, the RW model employs a softmax policy for action selection, with an inverse temperature parameter that transforms values into probabilities and a perseveration parameter that increases the weight of previously chosen actions. For full model details, see Appendix \ref{sec:learning}.

\vspace{+0.10cm}

\begin{figure*}[h]
    \centering
    \includegraphics[width=\textwidth, keepaspectratio]{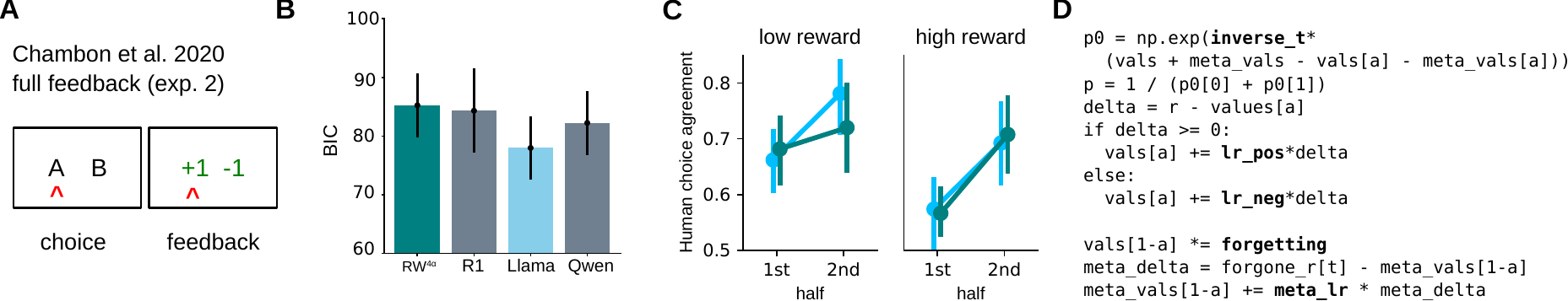}
    \caption{Experiment 2: Learning. A) Schematic of the learning task from \citealt{chambon2020information}, where participants chose between two options and received feedback for both, but only got the reward for the chosen option.
    B) Model fit comparison: LLM-generated models from Llama on average fit better than $RW^{4\alpha}$.
    C) Posterior predictive checks showing close alignment between human data and predictions of the best LLM model in both low and high reward blocks.
    D) Code of the best LLM-generated model (Llama), which displays asymmetric learning rates along with forgetting of values, and a dedicated fictive trace for counterfactual outcomes.}
    \label{fig:learning}
\end{figure*}

\paragraph{Results.}

We found that Llama (BIC$_{M \pm SEM}$ = 77.97 $\pm$ 5.40) produced a model that performed slightly better than $RW^{4\alpha}$ (BIC$_{M \pm SEM}$ =  85.24 $\pm$ 7.95); see Figure \ref{fig:learning}B. However, we found that, in terms of the exceedance probability (EXP), the LLM model (0.97) scored significantly higher compared to the best model reported in the literature (0.03). In addition to differentiating learning rates based on the valence of the prediction error, the best LLM-generated model included two key components: a forgetting parameter that decays the value of the unchosen action (potentially capturing working memory limits), and a separate meta‑value function (and learning rate) that learns uniquely from forgone outcomes (capturing regret/relief signals in a dedicated fictive trace); see Figure~\ref{fig:learning}D. For posterior predictive checks, we assessed the agreement between model-predicted choices and actual human choices separately for high- and low-reward blocks. The LLM-generated model matched human behavior as well as $RW^{4\alpha}$(Figure~\ref{fig:learning}C).

\paragraph{Scientific Insight.}
The GeCCo-generated model separates two sets of Q-values: one updated from chosen outcomes and another counterfactual set updated from forgone outcomes. Whereas standard models often merge these via asymmetric learning rates, GeCCo keeps them distinct, raising questions about how experienced and counterfactual values are weighted in decision-making. The model also applies forgetting to unchosen actions, creating temporal asymmetry in value retention and eliminating the need for an explicit choice stickiness parameter—often included mainly for model fit rather than theory. This illustrates how GeCCo can inspire rethinking and improving cognitive models.


\section{Experiment 3: Planning}

\paragraph{Task.}
A widely used paradigm in the planning literature is the two-stage task introduced by \cite{daw2011}. The goal of this task is to investigate whether people learn action-reward contingencies using habitual, reward-driven learning (referred to as model-free), or they make decisions based on the underlying state-transition structure that they estimate from experience (model-based). We considered a planning dataset from \cite{feher2020humans}, which replicated the original study \citep{daw2011}; see Appendix \ref{app:human_experiments} for details of the experiment. 


\paragraph{Baseline cognitive model.}
For baseline, we considered the Hybrid model from \cite{daw2011} - the best performing model from the literature, which effectively combines model-free action values from a SARSA($\lambda$) algorithm with model-based values (updated based on the transition probabilities), weighted by a free parameter. For detailed equations, see Appendix \ref{sec:planning}.

\begin{figure*}[h]
    \centering
    \includegraphics[width=\textwidth, keepaspectratio]{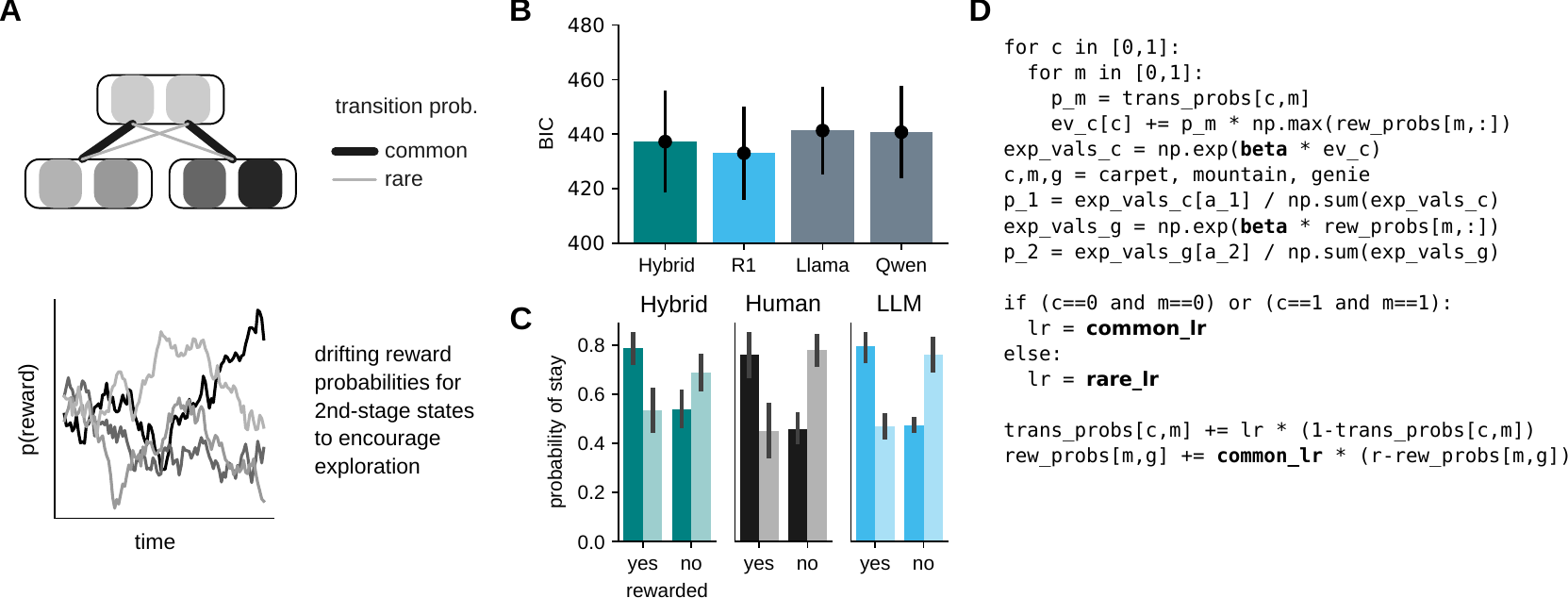}
    \caption{Experiment 3: Planning. A) Schematic of the planning task from \cite{feher2020humans}, where participants took two steps in a stochastic environment with common and rare transitions and fluctuating rewards.
    B) Model fit comparison: the model generated by R1 had a lower average BIC score compared to the Hybrid model.
    C) Posterior predictive checks showing the same pattern across humans, literature model and R1-generated model of repetition of common (dark) and rare (bright) transitions depending on if the previous action was rewarded.
    D) Code of the best LLM-generated model (R1), which uses separate learning rates for common and rare transitions and inverse temperature for exploration -- omitting discounting of rewards.}
    \label{fig:planning}
\end{figure*}

\paragraph{Results.}
As shown in Figure \ref{fig:planning}B, R1 generated the best overall model that achieved a lower mean BIC (432.47 $\pm$ 19.31) than the hybrid (437.38 $\pm$ 18.72). While the BIC difference was not significant (t-test (Hybrid > R1 BIC): t(15) = 0.61; p = 0.27), the exceedance probability (EXP) favored the R1 model (R1 EXP = 0.85 versus Hybrid EXP = 0.15). Posterior predictive checks revealed that the LLM-generated model, like humans, made action selections consistent with a more model-based strategy; see Figure \ref{fig:planning}C.  The code generated by R1 describes a version of the transition dependent learning rate model that uses two separate learning rates for common and rare transitions, respectively (similar to one of the candidate models reported by \cite{feher2020humans}). Furthermore, it learns the transition probabilities, indicating that it is not model-free and behaves similarly to humans. It also incorporates an inverse temperature parameter to regulate exploration, but notably omits a discounting parameter, which is typically included in multi-step planning tasks to devalue distant rewards. Drawing on reinforcement learning literature, the model implements value iteration \citep{bellmann1958}, taking the maximum over future value estimates (line 4 of the code) to update the expected value for the first-stage actions; see Figure \ref{fig:planning}D.

\paragraph{Scientific Insight.}
The GeCCo-generated model offers a parsimonious account of planning, showing how behavior based on learned transition structures can emerge from general principles of transition learning and value updating, without explicitly separating model-based and model-free systems. This provides an alternative to traditional reinforcement learning frameworks that rely on multiple canonical components and aligns with recent calls to move beyond the strict MB/MF dichotomy \citep{collins2020beyond}.


\section{Experiment 4: Working memory}
\paragraph{Task.}
The reinforcement learning–working memory (RL-WM) paradigm was designed to disentangle reinforcement learning (RL) and working memory (WM) contributions to the learning process \citep{collins2012much}. In this paradigm, participants learn a varying number of stimulus–action (S-A) associations per block via feedback, receiving a reward for correct responses to individual stimuli. Varying the cognitive load targets the WM capacity by increasing the number of associations and the delay between stimulus repetitions. We used data from the version reported in \cite{rmus2023age} with two set sizes: three (low) and six (high) S-As ; see Appendix \ref{app:human_experiments} for details.

\paragraph{Baseline cognitive model.}
As a baseline, we used the RL-WM model, which consists of independent but interacting reinforcement learning (RL) and working memory (WM) modules to isolate their respective contributions to learning. The RL module captures incremental, capacity-unlimited learning, whereas the WM module supports one-shot learning with perfect retention from the previous trial but is capacity-limited and subject to decay. For detailed equations, see Appendix \ref{sec:working_memory}.

\vspace{+0.10cm}
\begin{figure*}[h]
    \centering
    \includegraphics[width=\textwidth, keepaspectratio]{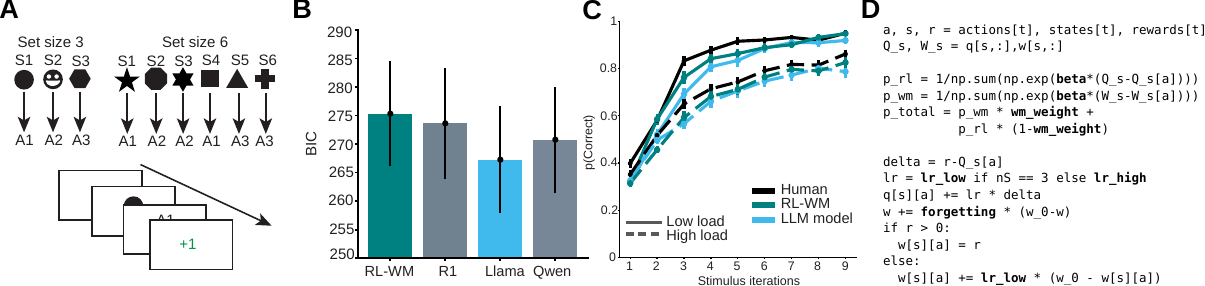}
    \caption{Experiment 4: Working Memory. A) Schematic of the reinforcement learning - working memory task from \cite{rmus2023age}, where participants learned a varying number of state-action associations. B) Model fit comparison: Llama-generated models outperformed the best literature model. C) Posterior predictive checks showing close alignment between human data and predictions of the best LLM model. D) Code of the best LLM-generated model (Llama), which distinguishes between fast learning under low cognitive load (working memory) and slower learning under high cognitive load.}
    \label{fig:working_memory}
\end{figure*}

\paragraph{Results.}

We found that Llama (BIC$_{M \pm SEM}$ = 267.25 $\pm$ 9.31) produced a model that, on average, had a lower average BIC score than the RL-WM model (BIC$_{M \pm SEM}$= 275.2 $\pm$ 9.21; t-test (RL-WM > Llama BIC): t(24) = 14.2; p $<.01$); Figure~\ref{fig:working_memory}B. In terms of exceedance probability (EXP), the Llama-generated model (0.99) scored higher than RL-WM (0.01). The LLM-generated cognitive model implemented differences in learning based on low vs. high cognitive load, but did so in the RL module, rather than using the cognitive load-dependent WM weight (as is the case in the original RL-WM model); Figure~\ref{fig:working_memory}D. 
Posterior predictive checks revealed that the LLM-generated model captured key patterns in human behavior reasonably well: faster, asymptotic learning in the low cognitive load condition typically associated with WM, and incremental, RL-like learning in the high cognitive load condition; Figure~\ref{fig:working_memory}C.

\paragraph{Scientific Insight.}
The LLM-generated model captured behavioral patterns and fit the human data well, even though the cognitive load manipulation primarily affected the RL rather than the WM component \citep{collins2012much}. This raises questions about where cognitive load effects arise and challenges the assumption that RL processes are relatively insensitive to load. Notably, GeCCo’s proposal aligns with evidence that RPE signals—central to reinforcement learning—are blunted under low cognitive load \citep{collins2017working}, suggesting that the RL system may adjust learning rates based on WM load.


\section{Control experiments}

\paragraph{Contribution of different model features.} To examine the importance of LLM features—reasoning ability, base model family, and model size—we fit a mixed-effects regression predicting the BIC of the best LLM-generated models from these three factors, with participant-specific random intercepts. Reasoning ability (as in R1) yielded slightly better models ($\beta_{M \pm \text{SEM}}$ = -0.266 $\pm$ 1.029, $p$ = 0.07), though the effect was not significant. In contrast, Llama outperformed Qwen as a base model ($\beta_{M \pm \text{SEM}}$ = 5.624 $\pm$ 1.029, $p$ = 0.03), despite Qwen having more parameters—likely due to differences in training data and/or training protocols. 

\paragraph{Causal role of different prompt components.} 
We conducted an ablation study to assess the causal contribution of each prompt component—feedback, code template, task description, and participant data sequences—by systematically ablating each element and rerunning the model generation pipeline (see Figure~\ref{llm-prompt-structure} for the full prompt structure). To estimate their joint influence, we fit a linear mixed-effects regression predicting post-ablation BIC scores from binary-coded prompt components, with random intercepts per participant. Feedback was the strongest predictor of model performance ($\beta_{M \pm \text{SEM}}$ = -17.040~$\pm$~2.287, $p$ < 0.05), followed by participant data ($\beta_{M \pm \text{SEM}}$ = -12.836~$\pm$~2.287, $p$ < 0.05) and task description ($\beta_{M \pm \text{SEM}}$ = -9.834~$\pm$~2.287, $p$ < 0.05). The code template had the weakest effect ($\beta_{M \pm \text{SEM}}$ = -9.359~$\pm$~2.287, $p$ < 0.05; see Figure~\ref{fig:ablation}).

\paragraph{Data contamination analysis.} 
We checked for data leakage in the prompts used to generate cognitive models using the LogProber method \citep{yax2024assessing}. 
Upon applying it to Llama (the base model that generated the best-performing cognitive models most frequently), we found no evidence of data contamination, with the acceleration term substantially below the threshold of 1 recommended by \cite{yax2024assessing} for the four human experiments. Specifically, it was 0.0171 for the prompts used for decision-making, 0.1147 (full feedback) and 0.1409 (partial feedback) for learning, 0.1718 for planning, and 0.6481 for working memory; see Figure \ref{fig:data_contamination} for details.
\paragraph{Recovering ground truth models in synthetic datasets.} 
Synthetic datasets generated through model simulations provide full control over task behavior and access to ground truth, making them ideal for benchmarking. As a sanity check, we evaluated our approach on synthetic data from two domains: learning and decision making. In both cases, the LLM inferred models that closely matched the true data-generating processes, supporting the validity of our method. In decision making, the LLM recovered key heuristics: single-feature focus for Take the Best and cross-feature comparisons for Tallying. In learning, it correctly identified the Rescorla-Wagner (RW) model with two learning rates when fitted to corresponding data, and for data from an RW model with stickiness, it produced a similar value-decay formulation. See Appendix \ref{sanity_check:decision_making}, Appendix \ref{sanity_check:learning}, Figure \ref{llm-heuristics}, and Figure \ref{llm-bandit} for details.
\paragraph{Explainable variance captured in human behavior.} 
To assess explainable variance in human behavior captured by LLM-generated models, we compared its negative log-likelihood to that of \textsc{Centaur} \citep{binz2024centaur} - a foundation model trained to predict human behavior across multiple cognitive domains - on held-out data. As \textsc{Centaur} outperforms domain-specific models, it serves as a proxy for maximum explainable variance. The LLM-generated models matched or exceeded its performance across domains (see Table~\ref{tab:llk_comparison}), indicating that they captured most of the explainable variance while remaining interpretable.

\begin{table}[!h]
\centering
\caption{Mean negative log‐likelihoods (M$\pm$SE) and paired $t$‐test results of \textsc{Centaur} vs.\ LLM-generated cognitive model across four task domains.}
\label{tab:llk_comparison}
\vskip 0.14in
\begin{small}
        \begin{tabular}{lllll}
        \toprule
        \textsc{\textbf{Domain}} & \textsc{\textbf{Centaur} (M$\pm$SE)} & \textsc{\textbf{LLM} (M$\pm$SE)} & \textsc{\textbf{$t$}} & \textsc{\textbf{$p$}} \\
        \midrule
        Decision‐making  & $15.41 \pm 1.99$    & $\textbf{15.08} \pm \textbf{2.29}$    & $0.27$   & $0.78$    \\
        Learning         & $53.58 \pm 3.31$    & $\textbf{26.17} \pm \textbf{2.79}$    & $18.83$  & $<0.001$  \\
        Planning         & $\textbf{206.00} \pm \textbf{11.14}$  & $214.10 \pm 10.59$  & $-1.42$  & $0.18$   \\
        Memory           & $127.44 \pm 4.68$   & $\textbf{120.67} \pm \textbf{4.68}$   & $5.75$  & $<0.001$  \\
        \bottomrule
        \end{tabular}
\end{small}
\vskip -0.1in
\end{table}

\section{Discussion}

We developed GeCCo, a novel pipeline that uses open-source LLMs to generate cognitive models as Python functions and iteratively refine them via feedback. We applied GeCCo to paradigms from four cognitive domains and found that LLM-generated models matched or outperformed domain-specific baselines in all cases. Control experiments showed that 1) the base LLM model family was an important determinant of performance, 2) all prompt components contributed meaningfully, but iterative feedback was causally the most important, and 3) No significant data leakage occurred in task prompts. We also demonstrated that GeCCo recovered ground truth models from simulated data and produced interpretable models that outperformed \textsc{Centaur} in predictive power, indicating that LLM-generated models captured most of the explainable variance in human behavior.


\subsection{Related work}

\paragraph{Model discovery with LLMs.}
The long-standing goal of automated model discovery holds promise of accelerating and democratizing scientific discovery by reducing dependence on prior expertise. Until recently, such efforts relied on domain-specific languages and hand-crafted search algorithms to explore predefined model spaces \citep{kemp2008discovery, lloyd2014automatic, musslick2024automatingpracticescience, gulwani2011automating, steinruecken2019automatic, hewson2023bayesian}. 
Recent advances in LLMs have overcome many of these limitations. Researchers have used LLMs to automate statistical model discovery \citep{li2024automated}, solve classic ML tasks \citep{xiao2024verbalized}, support hypothesis generation and refinement \citep{zhou2024hypothesis}, propose valid domain rules (e.g., chemistry) \citep{zheng2023large}, and even automate the entire scientific pipeline, from experiment design to peer review, in machine learning \citep{lu2024ai}. In cognitive modeling, LLMs have been used to translate human strategy descriptions into executable code \citep{xie2024sorting}, and to generate cognitive models from task descriptions and data using evolutionary search \citep{castro2025discovering}. Our approach complements these concurrent efforts by offering a lightweight open-source method that relies purely on in-context learning, enabling much faster convergence (hours instead of days). 


\paragraph{Code writing abilities of LLMs.}
The promise of LLMs in model search lies not only in their broad knowledge but also in their ability to interpret natural language and synthesize executable code. Using LLMs to generate models in general-purpose languages like Python provides a path beyond handcrafted, domain-specific languages for automating model discovery \citep{austin2021program, ni2024l2ceval}. Prior work has shown that LLMs excel at mathematical and programming tasks \citep{austin2021program, ni2024l2ceval, perez2021automatic} and can generate mathematical functions and probabilistic Python programs that model input data \citep{li2024automated, xiao2024verbalized, zheng2023large}. More recently, LLMs have advanced program induction, with successes in domains such as abstract reasoning \citep{li2024arc} and graphical function induction \citep{li2024programmingByExample}.
\subsection{Discussion and Limitations}

A key advantage of our approach is that all results were obtained purely in-context using open-source LLMs, without any fine-tuning. This drastically reduces the barrier to entry, as researchers could use our pipeline off-the-shelf to generate alternative models for their behavioral data. GeCCo implements a hybrid optimization loop, where the LLM generates candidate models that are fit to the data offline via traditional optimization tools. This ensures that model selection remains data-driven, with the LLM acting as a proposal engine rather than an unverified model generator. 
Finally, our pipeline generalizes well across diverse task domains, making it adaptable to other researchers. These advantages suggest that our approach can help cognitive scientists explore a broader model space more efficiently.

Despite its success, our approach has some limitations. 
Our approach uses a simple domain-specific model as a template in the prompt. Although this could bias the LLM toward a particular model class, an ablation analysis showed that removing the template had the smallest impact on downstream model performance. To further reduce dependency on handcrafted templates, we also generated the template with the LLM itself, which produced models that fit the data just as well (see Appendix \ref{llm_generated_template}). 
Currently, our approach uses offline-computed model fit metrics (e.g., BIC) to generate feedback, consisting of the current best-performing model and previously tested parameter combinations. This could be enhanced by incorporating feedback from another LLM based on predefined criteria such as theoretical plausibility, parsimony, alternative hypotheses, or code quality, which may further reduce the need for human oversight. It is also important to note that the domains examined in this paper do not represent the full scope of cognitive modeling. Future work should extend our approach to other areas such as language processing and perception. Applying GeCCo to higher-dimensional data (e.g., vision or language) and generating models with many parameters may be challenging, but it represents an exciting direction for future research. Furthermore, to make our approach valuable beyond cognitive science, it should be evaluated on more naturalistic datasets that capture human behavior in real-world settings.


Overall, our findings suggest that LLMs can significantly advance computational modeling in cognitive science by democratizing access to complex model discovery and accelerating the pace of research. 

\section*{Acknowledgements}
We thank all reviewers for their constructive and thoughtful feedback. We also thank the authors of \citet{chambon2020information}, \citet{rmus2023age}, \citet{feher2020humans}, and \citet{hilbig2014generalized} for making the data from their study available. Furthermore, we thank the members of the ``Human-Centered Artificial Intelligence'' (HCAI Lab) for their comments, discussions, and support. This work was supported by the Volkswagen Foundation, Princeton AI Lab, Helmholtz Munich and the Deutsche Forschungsgemeinschaft (DFG, German Research Foundation) under Germany’s Excellence Strategy–EXC2064/1–390727645.15/18.

\bibliographystyle{apalike}
\bibliography{neurips_2025}

\newpage
\appendix

\definecolor{ceruleanblue}{rgb}{0.16, 0.32, 0.75}

\section{Prompt structure}

\begin{figure*}[hbt!]
    \centering
    \includegraphics[width=0.95\textwidth]{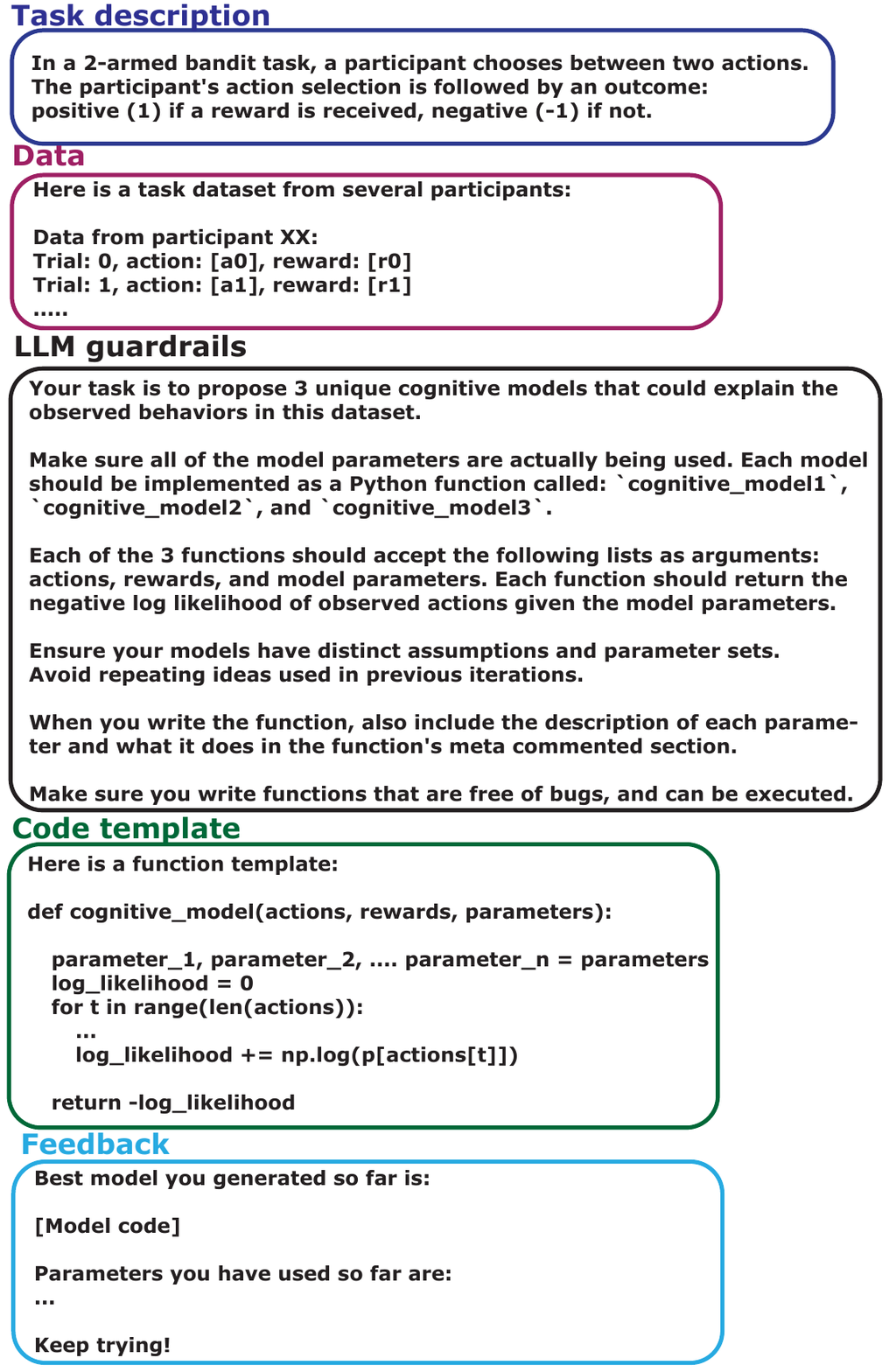}
    \caption{General structure of the LLM prompt we used across all of the domains. The prompt generally consisted of the paradigm description, data in the text format, Python function specifications (including the name of the function, required arguments and output), code function template, and the feedback (on iterations after the first one) established following offline evaluation of LLM-generated models. }
    \label{llm-prompt-structure}
\end{figure*}

\section{Prompts \label{all_prompts}}

\subsection{Decision making}

\begin{custombox_decision_making}
\label{prompt:decision_making}
This is a multi-attribute decision-making task, where participants have to choose the superior product between two options, labeled A and B.
Each option represented a fictitious product and they had to infer which product was superior in terms of quality in every trial.
For each option, they were provided with four expert ratings (with 1 representing a positive and 0 representing a negative rating).
The four experts differ in their validity.
The ratings of experts were given in descending order of their validity (having validities of 0.9, 0.8, 0.7, and 0.6 respectively).
Participants selected a product by selecting the corresponding option, i.e. A or B. Here is a task data set from several participants: \\

Data from participant 1:\\

Trial 1: Product A ratings: [1 1 1 1]. Product B ratings: [0 0 1 1]. Chosen option: A

Trial 2: Product A ratings: [1 0 0 0]. Product B ratings: [0 0 0 1]. Chosen option: A

Trial 3: Product A ratings: [0 1 1 1]. Product B ratings: [1 1 1 0]. Chosen option: B.

...

Propose three unique cognitive models that could explain the observed behavior in the dataset.

Each model should have distinct assumptions and parameters. Avoid repeating ideas used in previous iterations. Think step by step: How do participants use features or expert validites to make their decisions?

Each model should be implemented as a Python function named \lstinline{`cognitive_model1`}, \lstinline{`cognitive_model2`}, and \lstinline{`cognitive_model3`}. Each function should take following inputs: a list of choices, a list of option A feature lists, a list of option B feature lists and a list of model parameters. Each function should return the negative log likelihood of the observed choices given its parameters.

Clearly define each model parameter and explain its role in the function's commented section. All parameters (except inverse temperature) should have values between 0 and 1. Ensure the equations do not result in nonsense values (e.g., avoid division by zero). Make sure your Python functions are executable and bug-free.

Consider the following code as a function template:

\begin{verbatim}
  
def cognitive_model(choices, optionAs, optionBs, validities, params):

    w1, w3, w3, w4, temperature = params
    v1, v2, v3, v4 = validities

    log_likelihood = 0
    for t in range(len(choices)):
        option_A, option_B = optionAs[t], optionBs[t]
        value_A = np.array(option_A)
        value_B = np.array(option_B)
        scale_value_difference = temperature * np.sum(value_B - value_A)
        choice_probability_B = 1.0 / 
                (1.0 + np.exp(-scale_value_difference))

        # compute log probability for actual choice
        p = choices[t] * np.log(choice_probability_B) + 
        (1 - choices[t]) * np.log(1 - choice_probability_B)
        log_likelihood += p

    return -log_likelihood
\end{verbatim}

Your function:

\end{custombox_decision_making}

\subsection{Learning}

\begin{custombox_learning}
\label{prompt:learning}

This is a multi-armed bandit reinforcement learning task, where participants were asked to choose one of the two actions to gain monetary rewards. The task included two possible actions, and on each trial participants had to choose between the two.

They observed the reward for both their chosen and unchosen actions. Each action led to a reward according to an underlying probability distribution, which remained the same within a block but could change between blocks.

Here is a task data set from several participants:

Data from participant 1:\\
Block: 1, Trial: 1, Chosen action: 1, Reward for the chosen action: -1, Reward for the unchosen action: 1

Block: 1, Trial: 2, Chosen action: 1, Reward for the chosen action: -1, Reward for the unchosen action: -1

Block: 1, Trial: 3, Chosen action: 1, Reward for the chosen action: -1, Reward for the unchosen action: -1 

 ...\\

Your task is to propose 3 unique cognitive models that could explain the observed behaviors in this dataset. When generating the models think in steps - for example: if on trial t participant chose a specific action and observed a given feedback for the chosen action AND the non-chosen action both, what is their subsequent action choice?

Ensure your models have distinct assumptions and parameter sets. Avoid repeating ideas used in previous iterations.Make sure all of the model parameters are actually being used.

Each model should be implemented as a Python function called \lstinline{`cognitive_model1`}, \lstinline{`cognitive_model2`}, and \lstinline{`cognitive_model3`}.

Each of the 3 functions should accept the following lists as arguments: actions, rewards, forgone rewards, and model parameters. Each function should return the negative log likelihood of observed actions given the model parameters.\\

When you write the function, also include the description of each parameter and what it does in the function's meta commented section. \\

Make sure the equations do not lead to nonsense values (e.g. watch out for division by 0). Make sure you write functions that are free of bugs, and can be executed. Here is an initial model guess of how participants solve the task:
        
\begin{verbatim}
    def cognitive_model(actions, rewards, forgone_rewards, parameters):
    
    
    lr, neg_lr, inverse_temperature  = parameters
    values = np.array([0.5, 0.5])
    log_likelihood = 0
    for t in range(len(actions)):
        p0 = np.exp(inverse_temperature*(values-values[actions[t]]))
        p = 1/(p0[0]+p0[1])
        
        log_likelihood += np.log(p)

        delta = rewards[t] - values[actions[t]]
        if delta >= 0:
            values[actions[t]] = values[actions[t]] + (lr * delta)
        else:
            values[actions[t]] = values[actions[t]] + (neg_lr * delta)
            
    
    return -log_likelihood

\end{verbatim}

Your functions:

\end{custombox_learning}

\subsection{Planning}

\begin{custombox_planning}
\label{prompt:model_generation_planning}

In this task participants were instructed to obtain as many gold coins as possible by flying one of the two magic carpets (carpets {carpets[0]} and {carpets[1]}). After choosing one of the carpets, participants would fly to one of the two mountains (Pink or Blue) where they would encounter one of the two genies. Pink Mountain has genies {genies[0]} and {genies[1]}, and Blue Mountain has genies {genies[2]} and {genies[3]}.\\
Participants then chose one of the two encountered genies to ask for gold coins and was either given a gold coin or not. Different genies gave gold coins with varying probability. Therefore, this was a two-stage decision process: first, participant chose the flying carpet; second, after ending up on one of the mountains they chose the genie they would ask for gold coins.\\

In the task, the magic carpet {carpets[0]} generally flew to the Pink Mountain, and magic carpet {carpets[1]} generally flew to the Blue Mountain.
However, on rare occasions a strong wind would send the magic carpet to a less likely destination (e.g., magic carpet {carpets[0]} would end up flying to the Blue Mountain). The participants might leverage their knowledge of the transition structure in the task (e.g., which carpet likely goes to which mountain) to maximize the chance of encountering the genie that on average yielded more gold).\\

Here is a task dataset from several participants:

Participant 1:

Trial 0: The participant chose magic carpet A and ended up on the Blue Mountain.
The participant rubbed the lamp S and received 0 coins.\\
        
Trial 1:The participant chose magic carpet A and ended up on the Pink Mountain.
The participant rubbed the lamp W and received 1 coin.\\

...\\

Your task is to propose 3 unique cognitive models that could explain the observed behaviors in this dataset.
Think in steps - for example: if on trial t participant won a gold coin after choosing a magic carpet that took them to a mountain where they asked a genie for gold coins, what is their subsequent carpet choice? Do they consider how often they end up on that mountain after choosing that specific carpet? \\

Ensure your models have distinct assumptions and parameter sets. Avoid repeating ideas used in previous iterations.
Make sure all of the model parameters are actually being used.
Each model should be implemented as a Python function called \lstinline{`cognitive_model1`}, \lstinline{`cognitive_model2`}, and \lstinline{`cognitive_model3`}.\\

Each of the 3 functions should accept the following lists as arguments: action\_1, state, action\_2, reward, parameters. Each function should return the negative log likelihood of observed actions given its parameters.\\

When you write the function, also include the description of each parameter and what it does in the function's meta commented section. 
Note that for each parameter except the beta, the plausible bounds are between 0 and 1. Make sure the equations do not lead to nonsense values (e.g. watch out for division by 0).\\

Make sure you write functions that are free of bugs, and can be executed.
Here is an initial model guess of how participants solve the task:

\begin{verbatim}
def cognitive_model(action_1, state, action_2, reward, parameters):

    learning_rate, beta = model_parameters
    n_trials = len(action_1)

    transition_matrix =  np.array(
        [[.7, .3],
         [.3, .7]]
    )
    p_choice_1 = np.zeros(n_trials)
    p_choice_2 = np.zeros(n_trials)
    Q = np.zeros((3, 2)) 

    for trial in range(n_trials):
        max_q = np.max(Q[1:], axis=1)  
        q_stage1 = transition_matrix @ max_q

        # Compute probability for choice 1
        exp_values = np.exp(beta * q_stage1)
        p_choice_1[trial] = np.exp(beta * q_stage1[action_1[trial]])/
                            np.sum(exp_values)

        # Compute probability for choice 2
        state_idx = state[trial] + 1  # Ensure correct state indexing
        exp_values_mf = np.exp(beta * Q[state_idx, :])
        p_choice_2[trial] = np.exp(beta * Q[state_idx, action_2[trial]])/
                            np.sum(exp_values_mf)

        delta = reward[trial] - Q[state_idx, action_2[trial]]
        Q[state_idx, action_2[trial]] += learning_rate * delta


    eps = 1e-10
    log_loss = -(np.sum(np.log(p_choice_1 + eps)) 
            + np.sum(np.log(p_choice_2 + eps)))

    return log_loss

\end{verbatim}

Your functions:

\end{custombox_planning}

\subsection{Working memory}

\begin{custombox_wm}
\label{prompt:working_memory}

In this task participants are presented with a varying number of states (either 3 or 6), and asked to select one of the 3 possible actions. For each state there is a fixed correct action.\\

Following the action selection, participants observed feedback (0 or 1). Possible states reset at the start of each block (that is, there were no overlapping states between the blocks).\\

The objective of the experiment was to understand differences in learning across different cognitive load conditions (3 vs 6 state-action pairs).\\

Here is a task dataset from several participants: \\

Data from participant 1:

Block: 0, Set size:3, Trial: 0, State: 0, Chosen action: 0, Reward: 1
 
Block: 0, Set size:3, Trial: 1, State: 1, Chosen action: 1, Reward: 1
 
Block: 0, Set size:3, Trial: 2, State: 2, Chosen action: 2, Reward: 0
 
...\\

Ensure your models have distinct assumptions and parameter sets. Avoid repeating ideas used in previous iterations. Make sure all of the model parameters are actually being used. Each model should be implemented as a Python function called \lstinline{`cognitive_model1`}, \lstinline{`cognitive_model2`}, and \lstinline{`cognitive_model3`}. Each of the 3 functions should accept the following arrays as arguments: states, actions, rewards, blocks and model parameters. Each function should return the negative log likelihood of observed actions given its parameters.\\

When you write the function, also include the description of each parameter and what it does in the function's meta commented section. Note that for each parameter except the inverse temperature, the plausible bounds are between 0 and 1. Make sure the equations do not lead to nonsense values (e.g. watch out for division by 0).\\

Make sure you write functions that are free of bugs, and can be executed.
Here is an initial model guess of how participants solve the task:

\begin{verbatim}
def cognitive_model(states, actions, rewards, blocks, parameters):
        lr, wm_weight, softmax_beta = parameters
        softmax_betawm = 50
        nA = 3
        nS = len(np.unique(block_states))
        q = (1 / nA) * np.ones((nS, nA))
        w = (1 / nA) * np.ones((nS, nA))
        w_0 = (1 / nA) * np.ones((nS, nA))
        log_p = 0
        for t in range(len(block_states)):
            a = block_actions[t]
            s = block_states[t]
            r = block_rewards[t]
            Q_s = q[s,:]
            W_s = w[s,:]
            p_rl = 1 / np.sum(np.exp(softmax_beta*(Q_s-Q_s[a])))
            p_wm = 1 / np.sum(np.exp(softmax_betawm * (W_s - W_s[a])))
            p_total = p_wm*wm_weight + (1-wm_weight)*p_rl
            log_p += np.log(p_total)
            delta = r-Q_s[a]
            q[s][a] += lr*delta
            if r > 0:
                w[s][a] = r
            else:
                w[s][a] += lr * (0-w[s][a])
    return -log_p

\end{verbatim}

Your function:

\end{custombox_wm}

\section{Decision making: Sanity checks using synthetic data \label{sanity_check:decision_making}}

\paragraph{Task.}
We designed a task in which decision making agents chose between two options (A and B). Each option is characterized by three features, represented as integers ranging from 0 to 100.

\paragraph{Heuristics.}
We specifically examined two: the Take the Best and the Tallying heuristic. The Take the Best heuristic selects an option based solely on a single prioritized feature, ignoring comparisons across other features \citep{gigerenzer1996reasoning}. Specifically, only the prioritized feature is used to evaluate the two options, with the option that has the higher value for the prioritized feature winning in the comparison.  
The Tallying heuristic instead compares the two options based on all three features, counting the number of features for which one option has a higher value than the other, and favoring the option that has a higher number of superior features.

\paragraph{Synthetic data generation.}
We generated 80 decision problems (each problem including two sets of three features), in accordance with the task specification. During task generation, we ensured that the number of times option A or B is superior is balanced. For Take the Best simulations we deliberately prioritized the second feature to avoid the LLM making a potentially misleading assumption that the first feature should be prioritized. We simulated 40 decisions based on each of the two heuristics. Importantly, we initially ensured that our examples were unambiguous -avoiding cases where both heuristics would lead to the same decision. This approach guaranteed the identifiability of decision patterns unique to each heuristic. To further test the robustness of our approach, we simulated a second data set that introduced noise to the data simulation. We did this by increasing the proportion of decisions in which the final output of the heuristic was flipped, resulting in the opposite choice than the heuristic would actually make. We considered three noise levels: 0, 0.25, and 0.5.  This progressively increased the level of confusion between the two heuristics (Figure \ref{llm-heuristics}B). At a noise level of 0.5, the decision patterns for both heuristics are indistinguishable and are equivalent to random guessing.

\paragraph{LLM prompting}

We queried the LLM to perform two tasks: (1) match the data to the source model (model identification) and (2) generate a cognitive model based on the observed data for both decision making and learning experiments (model generation). In the model identification task, the LLM prompt included the code for model candidates provided as Python strings along with simulated decisions. For the model generation task, the prompt included a description of the desired structure of the Python function (e.g., the function name, input arguments, and expected output). We considered three different prompting strategies: vanilla (containing only the description of the task setup), Step-Back \citep{zheng2023take}, and Chain of Thought (CoT; \citealt{wei2022chain}). This comparison aimed to identify the most effective prompting strategy for subsequent tests.
 
\subsection{Results}

\paragraph{Model identification.}

\begin{figure*}[hbt!]
    \centering
    \includegraphics[width=0.95\textwidth]{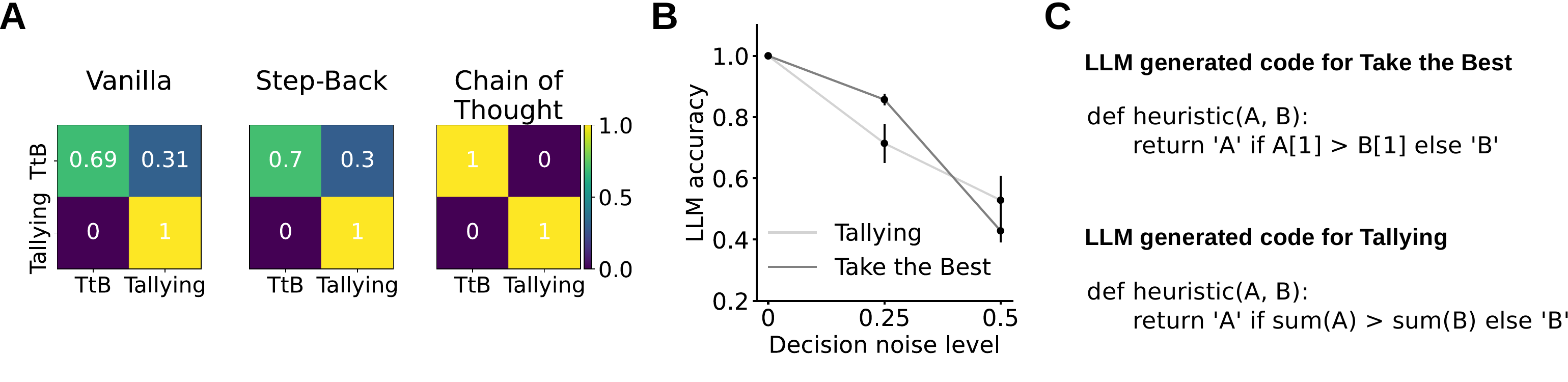}
    \caption{ A) Identifying the source decision making heuristic by using the LLM to relate data to the source simulation code. We prompted the LLM to identify which of the two heuristics (Take the Best or Tallying) underlies the behavioral data from the two-alternative forced choice task, where the agent chooses between two options defined by three features. While exploring the LLM's capacity to perform this task, we tried different prompting techniques. B) We tested the LLM with noisy decision making data, with injected noise increasing the confusion between the two heuristics. The LLM shows robustness to noise in the data - performance decreases proportionally with noise, but only reaches chance level when the heuristics are indeed indistinguishable (noise level of 0.5). Error bars represent standard error of the mean (SEM) across 10 runs. C) LLM-generated Python function heuristics closely align with the ground truth. The LLM-generated functions remained the same across 10 separate experiment runs.}
    \label{llm-heuristics}
\end{figure*}

Model identification enabled us to evaluate the LLM's ability to reason through decision making strategies and map the data to the underlying function. Consistent with previous results \citep{wei2022chain}, CoT prompting led to the best LLM performance (Figure \ref{llm-heuristics}A); therefore, we used CoT prompting for all subsequent experiments. We found that in the noise-free data set, the LLM was perfect at identifying the source model based on the data (Figure \ref{llm-heuristics}A). When evaluated on the noisy data set, the LLM robustly identified the ground truth heuristic at noise level of 0.25 (Take the Best mean accuracy: 0.86 (SEM = 0.02); Tallying mean accuracy: 0.71 (SEM = 0.06)). At a noise value of 0.5 - corresponding to random guessing in the data generating process (Figure \ref{llm-heuristics}B) - the LLM's heuristics predicted decisions at chance level (Take the Best accuracy $>$ 0.5 test : t(9)=1.80, p=0.10; Tallying accuracy $>$ 0.5: t(9)=0.27, p=0.79).

\paragraph{Model generation.}

Next, we tested whether the LLM could generate a decision making algorithm that aligned with the observed strategy patterns, simulated using either the Take the Best or Tallying heuristic. We evaluated the LLM-generated algorithms on unseen decision tasks, comparing their outputs to the predictions of the ground truth heuristic.

Our analysis of LLM-generated functions revealed that the LLM could successfully identify the two underlying heuristics: prioritizing a single feature for Take the Best simulations and performing across-feature comparisons for Tallying simulations (Figure \ref{llm-heuristics}C). The choices generated by the LLM-proposed models matched the ground truth heuristic choices with perfect accuracy in the evaluation tasks. It is notable, however, that for the Tallying heuristic there was a slight departure from the ground truth in the LLM-generated code -- using the total sum of features instead of a tally of superior features. There are corner cases where these strategies would make diverging predictions (e.g., if the feature values are not normalized / in the same range). 
Nevertheless, the LLM proposed an equally valid alternative to the true data generating process. 

To account for the noise in the noisy data set, the LLM-generated heuristics deviated more from the underlying heuristics. For noisy Take the Best data, the LLM still prioritized the second feature but modified the heuristic to apply only when a specific criterion (e.g., feature value differences above a certain threshold) was met. For noisy Tallying data, the LLM generated various strategies such as choosing based on the highest overall or minimum feature value. 


\section{Learning: Sanity checks using synthetic data \label{sanity_check:learning}}

\paragraph{Task.}
We implemented a two-armed bandit task, with each of the two options associated with a fixed probability of receiving a reward if selected (e.g. ${p(r = 1|a_1) = 0.20; p(r = 1|a_2) = 0.80}$). The rewards in our task were binary ($r \in \{0, 1\}$).

\paragraph{Learning models.}
The Rescorla-Wagner (RW) model \citep{rescorla1972theory} is commonly used to study learning dynamics in the bandit tasks. In the experiment, we considered the vanilla RW model and two variants of it - RW with two learning rates ($RW + \alpha^\pm$) and RW with stickiness ($RW + \kappa$).

The RW model posits that the value of each action ($V$) is determined by the history of rewards obtained from selecting that action. According to the RW model's learning rule, the value of the selected action $a$ ($V^a$) is updated on each trial \textit{t} as follows:

\begin{equation*}
{V^a_{t+1} = V^a_{t} + \alpha \cdot (r-V^a_{t})}
\end{equation*}

where $r-V^a$ is the reward prediction error - a learning signal that drives the adjustment of the selected action value, and $\alpha$ represents a learning rate that captures the extent to which the action value is modified by the prediction error.

Learning models commonly rely on the softmax policy in conjunction with the RW learning rule, providing a way to transform action values into probabilities. The softmax policy introduces the exploration parameter $\beta$, which controls the degree to which action selection is deterministic: 

\begin{equation*}
P(a) = \frac{\mathrm{exp}(\beta \cdot V^a_{t})}{\sum_{i=1}^{N} \mathrm{exp}(\beta \cdot  V^i_{t})}
\end{equation*}

\textit{The Rescorla-Wagner model with two learning rates} ($RW + \alpha^\pm$) differentiates between outcomes that are better/worse than expected. More precisely, the model uses two distinct learning rates for action value updating, contingent on the valence of the prediction error:

\begin{align*}
V_{t+1}^a &= \begin{cases}
V_{t}^a + \alpha^+\,(r-V^a_{t}) & \mbox{if } r-V^a_{t} \geq 0 \\
V_{t}^a + \alpha^-\,(r-V^a_{t}) & \mbox{if } r-V^a_{t} < 0
\end{cases} 
\end{align*}

\textit{The Rescorla-Wagner model with stickiness} ($RW + \kappa$) has the same learning rule as the vanilla RW but its policy differs in that the additional weight $\kappa$ is applied to the value of the action that was selected during the previous trial, resulting in a greater tendency to choose the previously selected action:

\begin{equation*}
    P(a) \propto \exp\left(\beta V + \kappa \mathbb{I}(a = a_{t-1})\right)
\end{equation*}

\paragraph{Synthetic data generation.}  To test how well we can recover the ground truth learning model, we simulated 100 agents from each of the two above-mentioned models on a two-armed bandit task, with each task comprising of 150 trials. The simulation parameters were randomly sampled for each agent in a range defined by plausible parameter bounds.

\begin{figure*}[htbp]
    \centering
    \includegraphics[width=.8\textwidth]{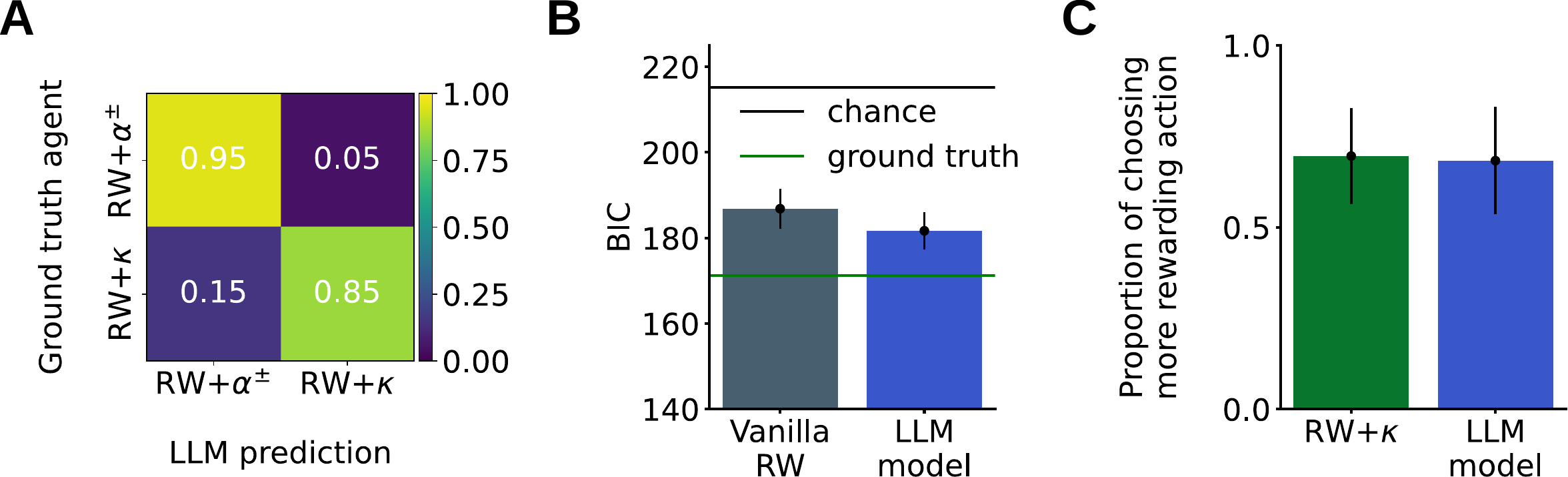}
    \caption{A) Model identification task. The LLM-generated 'ModelIdentification' function utilizes the SciPy differential evolution method to successfully differentiate between the two learning models. B) Evaluation of the LLM-generated cognitive model based on the data simulated from the $RW +\kappa$ revealed that it captured behavior better than the random guessing model and the vanilla RW. C) Simulation of the LLM-generated model showed that it captured the underlying propensity for choosing more rewarding actions. Error bars represent standard error of the mean (SEM) across simulated agents.}
    \label{llm-bandit}
\end{figure*}

\paragraph{LLM Prompting.}

Identifying the source model by reasoning through the long sequences of learning data, which consisted of 150 actions and rewards, is much more challenging than identifying the underlying decision making heuristic. Additionally, differently configured models can produce similar action/reward trajectories - a common challenge in models of bandit tasks \citep{wilson2019ten}. As a result, we modified the prompts we had developed for the decision making experiment.

\textit{Function generation for model identification.}  Unlike in the decision making case where the LLM directly returned the model identity, in the learning task the LLM instead generated a function for model identification. The function's arguments were predefined (e.g., lists of actions and rewards). The prompt encouraged the LLM to propose a method for matching the source model to the data without requiring step-by-step reasoning. The generated function was then manually evaluated to determine its accuracy in identifying the correct model.

\paragraph{Evaluation of LLM-generated cognitive models.} The best LLM-generated cognitive models were evaluated in two steps at the end of the final sampling run. First, we compared the Bayesian Information Criterion (BIC; \citealt{watanabe2013widely}) of the best model to the ground truth (or the best model from the literature for human data), random and a competing model.
Second, we manually implemented a simulation script based on the equations of the LLM-generated model. Using the best-fit parameters from the first step, the script simulated action choices according to the model's equations and parameters, with rewards determined by the probabilistic action–reward contingencies of the task. This step allowed us to assess how well the model captured behavioral patterns by comparing the simulated data with the ground truth.

\subsection{Results}

\paragraph{Model identification.}

We prompted the LLM to write a function, called \texttt{ModelIdentification}, that would identify the source model based on the underlying data in the learning experiments. We found that the LLM generated a function that performed an optimized search over possible model parameter values to find optimal log-likelihood, leveraging the differential evolution algorithm for optimization \citep{storn1997differential} from the SciPy library. The proposed function returned the identity of the model associated with the smaller negative log-likelihood. We executed this LLM-generated function offline to determine the identity of the model. As shown in Figure \ref{llm-bandit}A, we found that the $RW + \alpha ^\pm$ model can be successfully identified 95\% of times (SEM = 4) and the $RW + \kappa$ model could be identified about 85\% of times (SEM = 7). 


\paragraph{Model generation.}

For model generation, we run GeCCo for 10 iterations where the LLM generated three cognitive models that were fitted to the data offline on each run. The feedback was automatically constructed based on the model fits, and included as the part of the prompt in the subsequent iterations. Note that we only considered the best LLM-generated models across all runs for model fitting and comparison.  

We found that the LLM recovered the $RW + \alpha^\pm$ model correctly from its simulated data. That is, the best generated model was the RW model with two learning rates, based on positive and negative feedback (ground truth model BIC: 78.78 (SEM = 5.3), LLM-generated model BIC: 78.40 (SEM = 1.3). 

For the $RW + \kappa$ model, the LLM did not discover the ground truth model. Instead, the best-fitting model contained a value-decaying mechanism, which lowered the value of the non-selected action on each trial. This can be viewed as a way to model the forgetting, or the information decaying mechanism, in the learning process. The LLM-generated model fitted better than the random guessing model (Figure \ref{llm-bandit}B; $t(99) = 7.44, p < 0.001$) and the vanilla RW (Figure \ref{llm-bandit}B; $t(99) = 2.36, p = 0.01$). As a sanity check, we also compared the data simulated based on the LLM-generated model to the ground truth by quantifying the proportion of trials in which the simulated agent selected the more rewarding option (Figure \ref{llm-bandit}C). The results showed that the data simulated from the LLM-derived model approximated the ground truth data reasonably well (Figure \ref{llm-bandit}B, proportion of selecting the more rewarding action: ground truth: 0.69 (SD = 0.13); LLM-generated model: 0.68 (SD = 0.13). 

Additionally, for both data sets, we checked which cognitive model parameters the LLM proposed across other sampling runs, beyond quantitatively studying only the best model. We found that the parameters were reasonable and among those commonly cited in the cognitive modeling literature.

\begin{table}[!t]
\centering
\caption{Examples of parameters in LLM-proposed cognitive models across various sampling runs. Modeling of these mechanisms is documented in previous research \citep{wilson2019ten}.}
\label{tab:cognitive_model_params}
\vskip 0.14in
\begin{small}
        \begin{tabular}{l p{0.55\columnwidth}}
        \toprule
        \textsc{\textbf{Model Parameter}} & \textsc{\textbf{Explanation}} \\
        \midrule
        Decay & Forgetting mechanism \citep{paskewitz2022explaining} \\
        \midrule
        Random lapse & Random action-executions \citep{nassar2016taming} \\
        \midrule
        Bias & Preference for a particular action \citep{balcarras2016attentional} \\
        \midrule
        Dynamic scaler & Parameter (e.g., learning rate) adjustment based on the trial number \citep{diederen2015scaling} \\
        \midrule
        Exploration bonus & Directed exploration \citep{wilson2021balancing}  \\
        \bottomrule
        \end{tabular}
\end{small}
\vskip -0.1in
\end{table}

\begin{figure*}
    \centering
    \includegraphics[width=1.\textwidth, keepaspectratio]{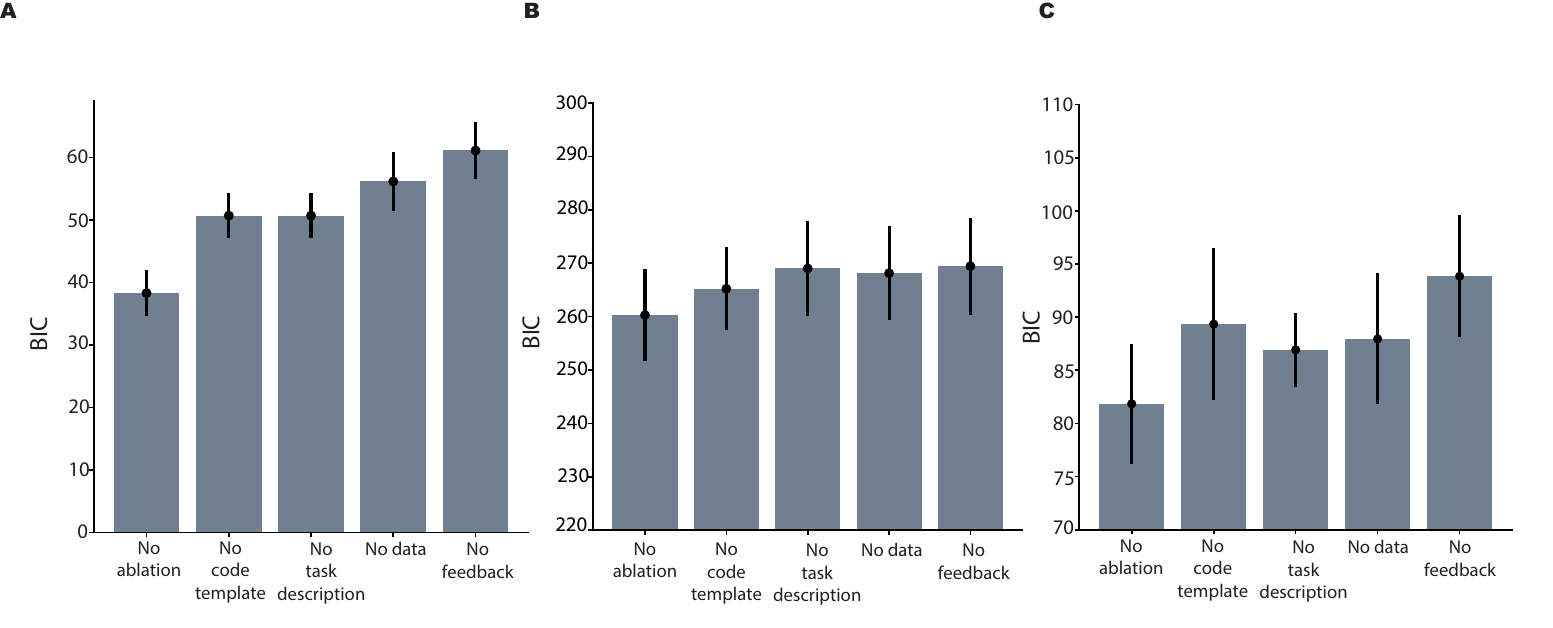}
    \caption{Ablation experiments. A) Ablation experiment in the decision making domain revealed that all components of the prompt were significant for contributing to performance of the LLM-generated models, with feedback removal from the prompt having the biggest effect. Results of the ablation experiment in the working memory domain B) and learning domain C) also support the impact of feedback.}
    \label{fig:ablation}
\end{figure*}


\begin{figure*}
    \centering
    \includegraphics[width=1.\textwidth, keepaspectratio]{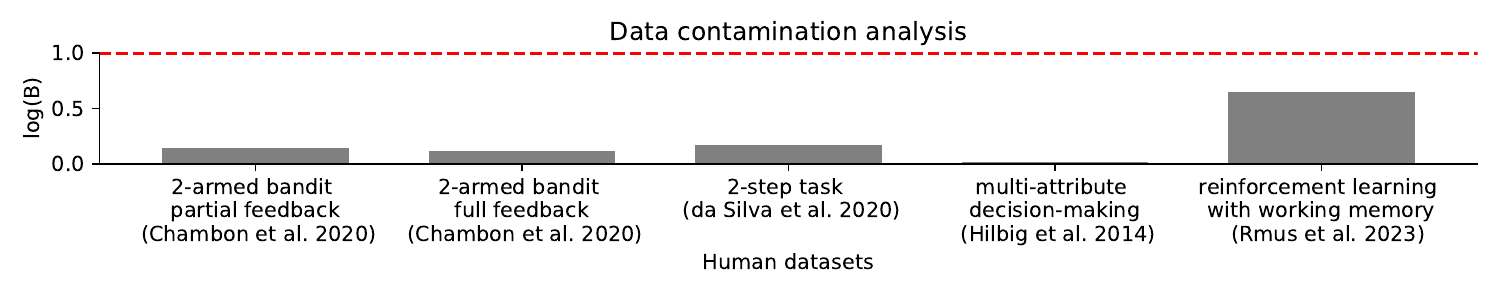}
    \caption{Data contamination analysis using the LogProber method (Yax et al. 2024).    LogProber fits a two-parameter exponential model to the cumulative log-likelihood of each sequence being checked for contamination: $f(x) = -A ( 1-exp^{-Bx})$. Prompts that are memorized from the pretraining data will show a high acceleration ($\log B$). Following the results presented in  (Yax et al., 2024), we set a threshold for possible contamination to $\log B \geq 1$.  We considered prompts up to the point of the choices of human participants, which contained the task instruction, instruction given to help with code generation, and the code template, for the four human datasets on the Llama-3.1-70B-Instruct (base model for all winning LLM). The analysis revealed no evidence of contamination, with the acceleration term being substantially below 1 for all four human experiments investigated. Specifically, it was 0.1409 for partial feedback condition \citep{chambon2020information}, 0.1147 for full-feedback condition \citep{chambon2020information}, 0.0171 for multi-attribute decision-making \citep{hilbig2014generalized}, 0.1718 for two-step task \citep{feher2020humans}, and 0.6481 for memory task \citep{rmus2023age}.}
    \label{fig:data_contamination}
\end{figure*}

\begin{figure*}[!t]
    \centering
    \includegraphics[width=1.\textwidth, keepaspectratio]{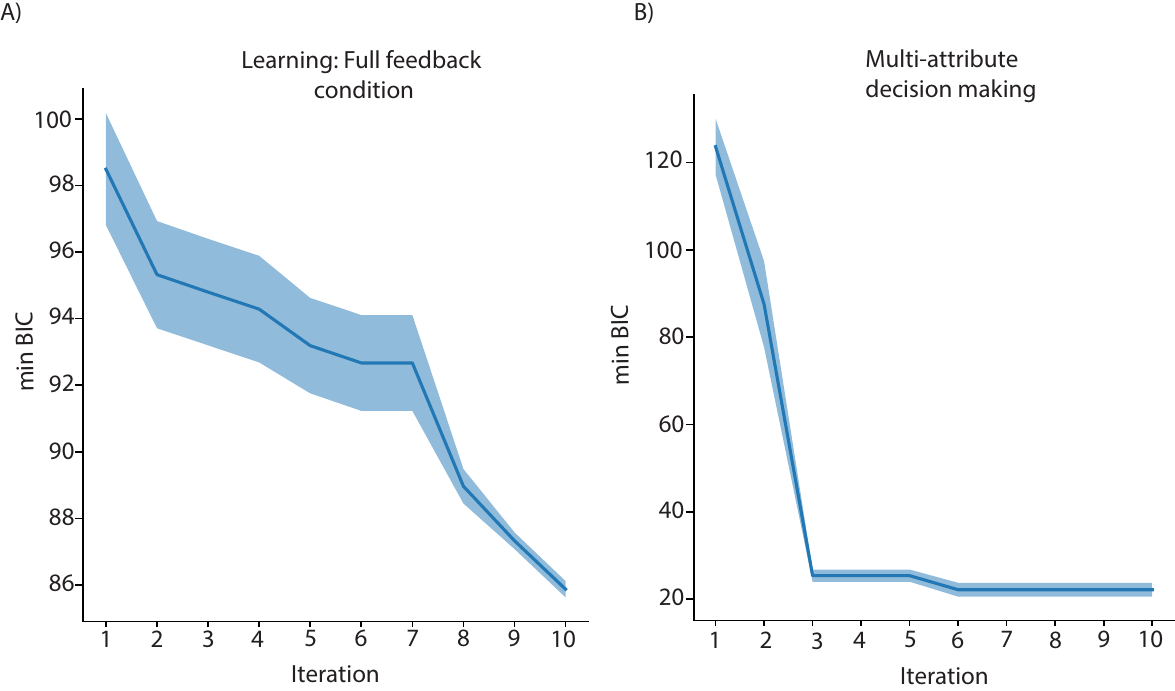}
    \caption{Minimum BIC scores decrease across sampling iterations. A) BIC improvement across iterations in the learning experiment. B) BIC improvement across iterations in the decision making experiment. Shaded area represents variability (SEM) across 5 different independent runs of GeCCo.}
    \label{fig:iteration_improvement}
\end{figure*}

\section{Human experiments} \label{app:human_experiments}

\subsection{Decision making}
In \cite{hilbig2014generalized}, 79 participants were instructed to repeatedly choose one of two options, labeled A and B, as shown in Figure \ref{fig:decision_making}A. They were told that these represent fictitious products and that they should infer which product is superior in terms of quality. In each trial, they were openly provided with the values of four binary features, explained to them as fictitious expert ratings. Furthermore, it was explained that the four ratings corresponded to ratings from four experts who differ in how well they typically predict product quality (i.e. feature validity), with their validities being .90, .80, .70, and .60, respectively. The participants performed 96 trials in total.

\subsection{Learning}
In the \cite{chambon2020information} study, 24 participants performed a two-armed bandit task designed to disentangle the effects of prediction-error valence on learning (see Appendix \ref{app:human_experiments}for additional study details). The task consisted of 16 blocks. The action reward probabilities varied across blocks and were sampled from high-probability reward values (0.9, 0.6) or low-probability reward values (0.4, 0.1). The rewards in this task were binary, with ($r \in \{-1, 1\}$). 
In half of the blocks, the participants received feedback solely based on their selected action (partial feedback condition; see Figure \ref{fig:partial_feedback}A). In the other half, the participants received feedback based on their selected action, as well as counterfactual feedback from the alternative action they did not select (full feedback condition; see Figure \ref{fig:learning}A). We focused on the counterfactual condition, as it presents a broader space of possible data explanations (e.g. differences in learning based on actual and forgone rewards). 

\subsection{Planning}
We used data from the “magic carpet” variant of the Daw two-step decision task, as implemented by \cite{feher2020humans}.In this version, participants were told they were embarking on a series of magic carpet rides. On each trial, they first chose between two magic carpets (first-stage choice), each of which probabilistically transported them to one of two mountains (second-stage states). Each mountain housed two genies, one of which participants could select to potentially receive a reward (second-stage choice). The reward probabilities associated with each genie drifted slowly over time, encouraging ongoing exploration.

The task structure is designed to differentiate between model-free and model-based learning strategies. First-stage transitions are probabilistic: a chosen carpet leads to its associated mountain with 70\% probability (common transition) and to the other mountain with 30\% probability (rare transition). Second-stage (genie) choices are deterministic.

A model-free learner reinforces actions that previously led to reward, regardless of the transition type. In contrast, a model-based planner takes into account the transition structure: if a reward was obtained following a rare transition, the optimal strategy would be to switch first-stage choices on the next trial to more reliably reach the same second-stage state via a common transition.

\subsection{Working memory}
Participants were instructed to learn stimulus-response associations in order to earn as much reward as possible \citep{rmus2023age}. The number of stimulus-action associations varied across different task blocks (3 or 6), with new stimuli being introduced in each block. Participants encounter one stimulus at the time on the screen, pressed a corresponding key and were subsequently provided with deterministic feedback. The rewards in this task were binary, with ($r \in \{0, 1\}$). Each stimulus was shown nine times within a block and stimulus order was pseudo-randomly shuffled.

\section{Cognitive models}

\subsection{Decision making: Take The Best (TTB), Equal weighting (EQW), Weighted sadditive hueristic (WADD) \label{sec:decision_making_cog_model}} 

Heuristics are simple, resource-efficient approaches that individuals use to navigate the decision making process effectively \citep{tversky1974judgment, gigerenzer1996reasoning}.
Assume $\mathbf x_o=(x_{o1},\dots,x_{oJ})$ to denote the feature vector for options $o\in\{A,B\}$, let $\mathbf w=(w_1,\dots,w_J)$ be the inferred feature weights and $\mathbf v=(v_1,\dots,v_J)$ be the feature validities, where $J$ is total number of features.  
As potential models \citep{gigerenzer1996reasoning}, Hilbig and colleagues considered three heuristics apart from the winning model (pWADD), namely, Take the Best (TTB), Equal Weighting (EQW), and Weighted Additive (WADD), to explain human choices.
The Take the Best (TTB) heuristic selects an option based solely on a single prioritized feature, ignoring comparisons between other features \citep{gigerenzer1996reasoning}. 
EQW heuristic instead compares options based on all four features, favoring the option that has a higher sum of superior features.
WADD heuristic is similar to EQW, but weighs the four features differently as per the provided feature validities. Finally, pWADD is the probabilistic variant of the WADD heuristic that uses inferred feature weights instead of using the feature validities (note $\beta$ is the temperature term). The equations for each of the models can be found in the following.

\begin{equation}
\begin{aligned}
\text{TTB}(A,B) &=
  \arg\max_{o\in\{A,B\}} x_{o\,j^{\dagger}}, 
  \qquad
  j^{\dagger}= \arg \max_{j \in \{1, 2, \ldots ,J\}} v_{j}
\\[10pt]
\text{EQW}(A,B) &=
  \begin{cases}
    A, & \displaystyle\sum_{j=1}^{J} (x_{Aj} - x_{Bj}) > 0,\\[6pt]
    B, & \text{otherwise}.
  \end{cases}
\\[12pt]
\text{WADD}(A,B) &=
  \begin{cases}
    A, & \displaystyle\sum_{j=1}^{J} v_j (x_{Aj} - x_{Bj}) > 0,\\[6pt]
    B, & \text{otherwise}.
  \end{cases}
\\[12pt]
\text{pWADD}(A,B) &=
  \begin{cases}
    p(A), & \displaystyle\frac{1}{1+\exp\!\bigl[-\beta\sum_{j=1}^{J} w_j (x_{Aj}-x_{Bj})\bigr]},\\[6pt]
    p(B), & 1 - p(A).
  \end{cases}
\end{aligned}
\end{equation}

\subsection{Learning: Rescorla Wagner model with 4 learning rates, softmax policy and perseveration \label{sec:learning}}

The winning model from \citealt{chambon2020information} is a variant of the RW model with four learning rates ( $\alpha^{c+},\alpha^{c-},\alpha^{u+},\alpha^{u-}$, $c=chosen, u=unchosen$). Thus, the model updates  values of actions differently, depending on whether the outcome of the action is better/worse than expected, as well as whether the action was chosen by the participant or not:

\noindent
\begin{minipage}[t]{0.42\textwidth}
\begin{equation*}
V_{t+1}^a = 
\begin{cases}
V_{t}^a + \alpha^{c+}\,(r-V^a_{t}) & \text{if } r - V^a_{t} \geq 0 \\
V_{t}^a + \alpha^{c-}\,(r-V^a_{t}) & \text{if } r - V^a_{t} < 0
\end{cases}
\end{equation*}
\end{minipage}
\hfill
\begin{minipage}[t]{0.46\textwidth}
\begin{equation*}
V_{t+1}^{1-a} = 
\begin{cases}
V_{t}^{1-a} + \alpha^{u+}\,(r'-V^{1-a}_{t}) & \text{if } r' - V^{1-a}_{t} \geq 0 \\
V_{t}^{1-a} + \alpha^{u-}\,(r'-V^{1-a}_{t}) & \text{if } r' - V^{1-a}_{t} < 0
\end{cases}
\end{equation*}
\end{minipage}

where \( r - V^a_t \) is the reward prediction error (RPE) for the chosen action, and \( r' - V^{1-a}_t \) is the RPE for the unchosen action based on the forgone reward. RPEs drive value updates, with \( \alpha^{c+} \) and \( \alpha^{u+} \) denoting learning rates for positive prediction errors of chosen and unchosen actions, respectively, and \( \alpha^{c-} \) and \( \alpha^{u-} \) for negative prediction errors.

The softmax policy introduces the exploration parameter $\beta$, which controls the degree to which action selection is deterministic:

\begin{equation*}
P(a) = \mathrm{softmax}(a) = \frac{\mathrm{exp}(\beta \cdot V^a_{t})}{\sum_{i=1}^{N} \mathrm{exp}(\beta \cdot  V^i_{t})}
\end{equation*}

We have also included the perseveration parameter $\kappa$ which assigns higher value to the action selected on the previous trial and thus captures the human tendency to repeat previously selected actions: \( P(a) \propto \exp\left(\beta V + \kappa\, \mathbb{I}(a = a_{t-1})\right) \).

\subsection{Planning: The Model-based/Model-free Hybrid model \label{sec:planning}}

The value of the selected first-stage action 
$a_1$ in state $s_1$ is updated based on the reward prediction errors from both the first and second stages, as follows:

\[
Q^{\text{MF}}_{t+1}(s_1, a_1) = Q^{\text{MF}}_t(s_1, a_1) + \alpha_1 \left[ Q^{\text{MF}}_t(s_2, a_2) - Q^{\text{MF}}_t(s_1, a_1) \right] + \alpha_1 \lambda \left[ r_t - Q^{\text{MF}}_t(s_2, a_2) \right]
\]

Model-free values of the second-stage action $a_2$ performed in the second-stage state $s_2$ are updated at the end of the trial (e.g., when the reward is observed) based on the reward prediction error—defined as the difference between the received reward $r_t$ and the current value of the chosen action:
\[
Q^{\text{MF}}_{t+1}(s_2, a_2) = Q^{\text{MF}}_t(s_2, a_2) + \alpha_2 \left[ r_t - Q^{\text{MF}}_t(s_2, a_2) \right]
\]

The model-based and model-free value of second-stage actions in state $s_2$ are equal:

\[
Q^{\text{MB}}_t(s_2, a_2) = Q^{\text{MF}}_t(s_2, a_2)
\]

The model-based value of each first-stage action is computed at the time of decision-making based on the values of the second-stage actions, as follows:

\[
Q^{\text{MB}}_t(s_1, a_1) = \sum_{s_2 \in \mathcal{S}} P(s_2 \mid s_1, a_1) \max_{a_2 \in \mathcal{A}} Q^{\text{MB}}_t(s_2, a_2)
\]

First-stage choices are determined by combining model-free and model-based value estimates for each state–action pair. These are integrated using a weighting parameter 
w (where $0 \leq w \leq 1$), which reflects the relative influence of model-based planning. The resulting action values are then passed through a softmax function to generate choice probabilities:

\[
P_t(s_1, a_1) = \frac{e^{\beta_1 \left[ w Q^{\text{MB}}_t(s_1, a_1) + (1 - w) Q^{\text{MF}}_t(s_1, a_1) + p \times \text{rep}_t(a_1) \right]}}{\sum_{a' \in \mathcal{A}} e^{\beta_1 \left[ w Q^{\text{MB}}_t(s_1, a') + (1 - w) Q^{\text{MF}}_t(s_1, a') + p \times \text{rep}_t(a') \right]}}
\]

\[
P_t(s_2, a_2) = \frac{e^{\beta_2 Q_t(s_2, a_2)}}{\sum_{a' \in \mathcal{A}} e^{\beta_2 Q_t(s_2, a')}}
\]

\subsection{Working memory: The Reinforcement Learning - Working Memory model \label{sec:working_memory}}

The learning dynamics in the RL module of this model are the same as in the RW model, see Appendix \ref{sec:learning}.

\begin{align*}
\delta_{\text{RL}} &= r - Q_{t}(s,a)\\
Q_{t+1}(s,a) &= \begin{cases}
Q_{t}(s,a) + \alpha^+\,\delta_{\text{RL}} & \mbox{if } \delta_{\text{RL}} > 0 \\
Q_{t}(s,a) + \alpha^-\,\delta_{\text{RL}} & \mbox{if } \delta_{\text{RL}} \leq 0
\end{cases} 
\end{align*}

where \textit{Q} denotes the action value, \textit{s} the presented stimulus and \textit{a} the chosen action.

The WM module, rather than integrating information over time, memorizes information from the previous trial. It also allows to encode WM stimulus action weights differently for positive and negative reward prediction error. In the case of positive reward prediction error,
\begin{align*}
W_{t+1}(s,a) &= r & \mbox{if } \delta_{\text{WM}} > 0\\
W_{t+1}(s,a) &= W_{t}(s,a) + v\,\delta_{\text{WM}} & \mbox{if } \delta_{\text{WM}} \leq 0
\end{align*}

where $v$ can be seen as the strength of the imperfection of the association and is identical to the relative neglect of negative feedback from the RL module $\frac{\alpha^-}{\alpha^+}$. Additionally, to model forgetting of the stimulus-action associations that are not seen in a current trial, the WM module has a decay parameter $\phi$ which pulls the WM weights to their initial values $W_0$.

\begin{equation*}
W_{t+1}(s_i, a_j) = W_t(s_i, a_j) + \phi \left( W_0(s_i, a_j) - W_t(s_i, a_j) \right) \quad \forall s_i\, \forall a_j \neq (s, a)
\end{equation*}

To make choices, the $Q$-values and WM weights are transformed into choice probabilities using a softmax policy:

\begin{align*}
P_{\text{RL}}(a|s) &= \frac{\exp(\beta^{RL} Q_t(s, a))}{\sum_{i=1}^{n_A} \exp(\beta^{RL} Q_t(s, a_i))} \\
P_{\text{WM}}(a|s) &= \frac{\exp(\beta^{WM} W_t(s, a))}{\sum_{i=1}^{n_A} \exp(\beta^{WM} W_t(s, a_i))}
\end{align*}

The RL and WM policies are then combined using a WM weight parameter $\omega$, determining the relative reliance on WM in taking an action. 

\begin{equation*}
P_{\text{RL-WM}}(a|s) = \omega^{ns} P_{\text{WM}}(a|s) + (1 - \omega^{ns}) P_{\text{RL}}(a|s)
\end{equation*}

$\omega$ is dependent on the set size of stimulus-response association pairs $n_s$, because as WM's capacity is exceeded with increasing set size, individuals should rely more on RL. The model thus predicts learning to be fast when set size is small due to high reliance on WM and learning to be more incremental when set size is high due to increased reliance on RL. Finally, the policy also captures value-independent random lapses:

\begin{equation*}
P = (1 - \epsilon)\, P_{\text{RLWM}} + \epsilon\, \frac{1}{n_A}
\end{equation*}

where $\frac{1}{n_A}$ denotes the uniform random policy and $\epsilon$ the noise parameter.

\section{Partial feedback (learning) experiment \label{partial_feedback}}

\begin{figure*}[!t]
    \centering
    \includegraphics[width=1.\textwidth, keepaspectratio]{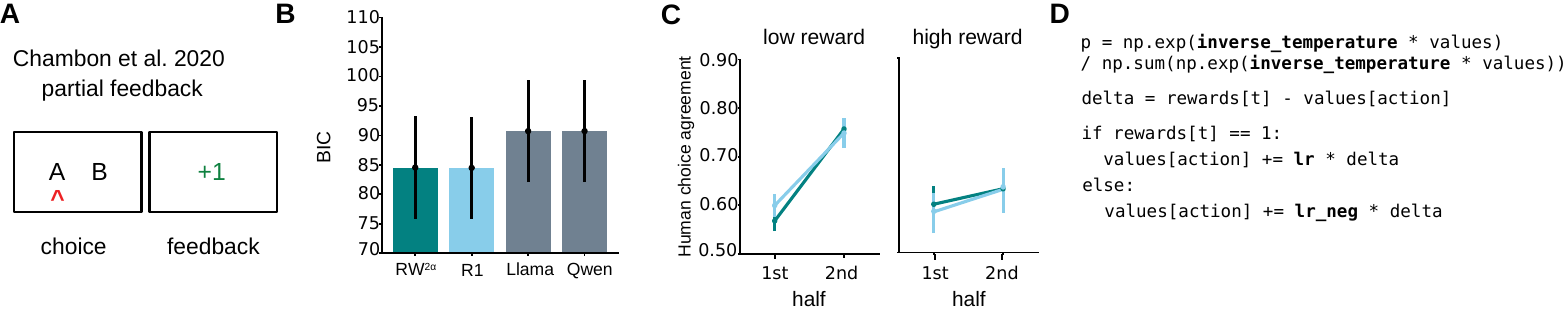}
    \caption{Learning experiment: partial feedback A) Schematic of the learning task from \cite{chambon2020information}, where participants chose between two options and received feedback only for the option they chose.
    B) Model fit comparison: LLM-generated model from R1 on average fit as well as $RW^{2\alpha}$.
    C) Posterior predictive checks showing close alignment between human data and predictions of the best LLM model.
    D) Code of the top LLM-generated model (R1), differentiating learning based on positive and negative feedback.}
    \label{fig:partial_feedback}
\end{figure*}

As a part of the learning experiment we considered two datasets by \cite{chambon2020information} - learning in partial feedback condition and full feedback condition. In the partial feedback condition participants only observed the outcome of the actions they chose (as opposed to also observing forgone rewards based on unchosen actions, Fig. \ref{fig:partial_feedback}A). \cite{chambon2020information} reported a two-learning rate RW model ($RW^{2\alpha}$) as the best model in their experiment. We applied the GeCCo to the partial feedback dataset using a vanilla RW model (with a learning rate and an inverse softmax temperature) as the function template. We found that the R1 proposed the best cognitive model (Fig. \ref{fig:partial_feedback}B) which closely aligned with $RW^{2\alpha}$ in terms of containing two learning rates, with a notable difference that the learning rates differentiated learning based on positive/negative feedback, rather than the prediction error valence; Fig. \ref{fig:partial_feedback}D. This model fits neatly into current debates about learning from valenced rewards, providing a valid alternative hypothesis to differentiating learning rate based on valence of reward prediction error.
Posterior predictive checks revealed close agreement between the LLM-generated model and human data; Fig. \ref{fig:partial_feedback}C.

\section{Posterior predictive checks \label{posterior_predictive_checks}}

Posterior predictive checks are a crucial tool for evaluating a model’s generative validity—that is, its ability to reproduce patterns observed in real behavior—thus offering insight into how well the model captures the underlying cognitive processes. To perform posterior predictive checks, we used ChatGPT to convert the LLM-generated model-fitting functions—originally designed to take participant behavior and parameter values as input and return negative log likelihood—into simulation functions that generate behavior from parameter values based on the model equations. Using the best-fitting parameters for each participant, we simulated behavior and compared it to the actual human data through domain-specific, informative analyses applied to both simulated and real datasets. 

In the decision making experiment, we computed the proportion of option choices that are explained by three different heuristics: Equal Weighting (EQW), Take The Best (TTB), and Weighted Additive (WADD); see section \ref{sec:decision_making_cog_model} for details about the heuristics. We conducted this analysis on 1) human data, 2) data simulated from the best LLM-generated model, and 3) data simulated from pWADD - the best performing model from the literature. We then examined correlation between the heuristic-specific proportions on a participant-level, as the data was simulated using the actual decision tasks (e.g. options and features) from the experiment participants experienced and best-fit model parameters for each participant. 

In the learning experiment, we simulated data for each participant using best fitting parameters, and block-specific reward probability contingencies specified in the \cite{chambon2020information} (high or low). We computed alignment with human choices for both the best performing LLM model, and $RW^{4\alpha}$ in high/low reward blocks separately, as well as for early/late trials in blocks to test whether the models capture temporal learning dynamics. 

In the planning experiment we performed the canonical analysis from \cite{daw2011}, which looks at the likelihood of participants repeating the same stage 1 response, contingent on whether 1) participants received reward at the end of the previous trial and 2) experienced a common or a rare transition en route to feedback. 

In the RL-WM experiment, we computed participant-specific learning curves showing the proportion of correct choices as a function of the stimulus iterations - the number of times participants encountered the stimulus. We used participant-specific stimulus sequences and correct stimulus-action mappings, to ensure that the behavior between simulated and real data is comparable.

\section{Using AIC for the best model selection in the GeCCo pipeline \label{aic_section}}

\begin{figure*}[h]
    \centering
    \includegraphics[width=1.\textwidth, keepaspectratio]{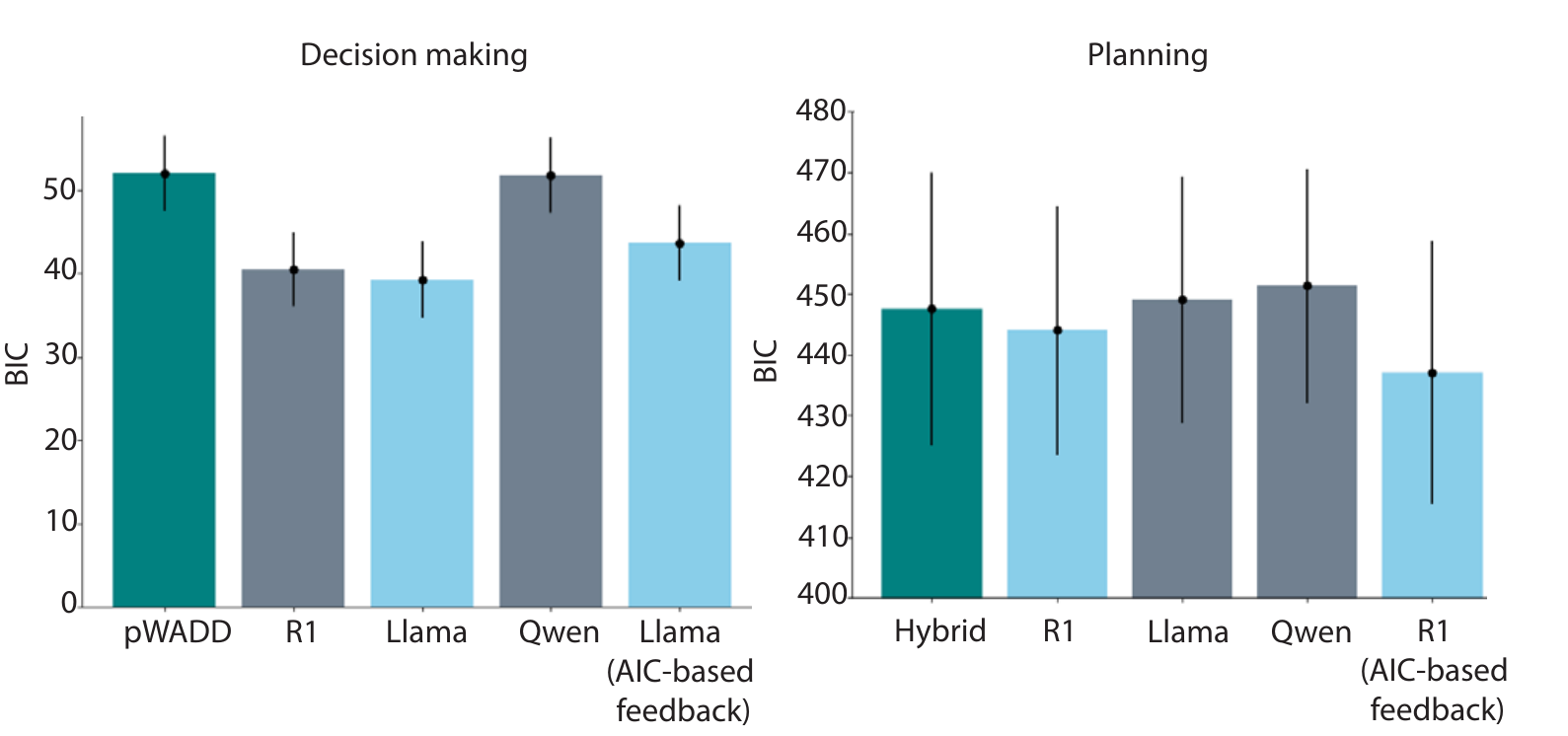}
    \caption{Comparison of the best GeCCo-generated models in the decision making and planning domains, including models from a pipeline variant using AIC-based feedback. This demonstrates that the pipeline is adaptable to different fit metrics and that AIC-based results closely replicate those obtained with BIC-based feedback.}
    \label{aic}
\end{figure*}

\section{Using LLM-generated template in the prompt\label{llm_generated_template}}

\begin{figure*}[h]
    \centering
    \includegraphics[width=1.\textwidth, keepaspectratio]{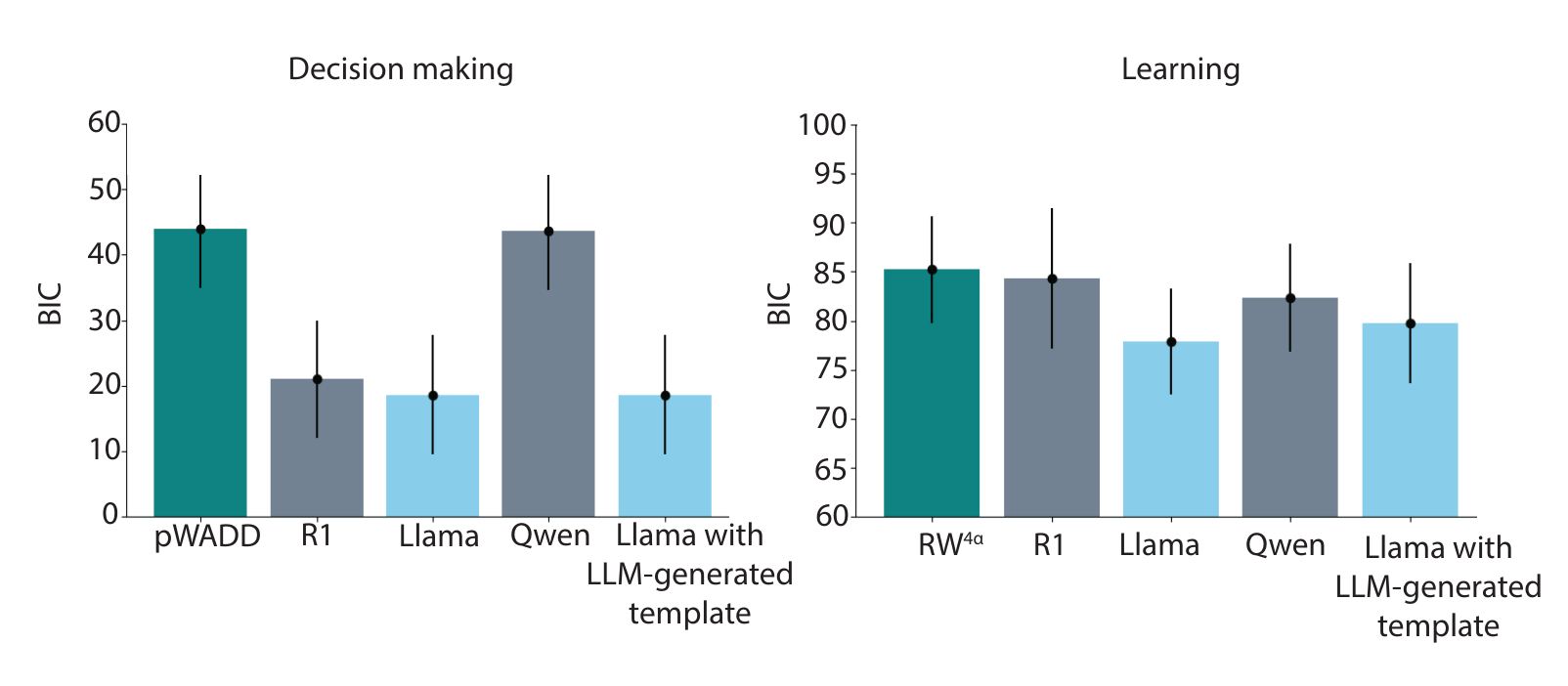}
    \caption{Comparison of the best GeCCo-generated models in the decision-making and learning domains, including models from a pipeline variant using an LLM-generated (rather than hand-crafted) template. The results show performance comparable to using a hand-crafted template.}
    \label{auto_template}
\end{figure*}


\newpage
\section*{NeurIPS Paper Checklist}

The checklist is designed to encourage best practices for responsible machine learning research, addressing issues of reproducibility, transparency, research ethics, and societal impact. Do not remove the checklist: {\bf The papers not including the checklist will be desk rejected.} The checklist should follow the references and follow the (optional) supplemental material.  The checklist does NOT count towards the page
limit. 

Please read the checklist guidelines carefully for information on how to answer these questions. For each question in the checklist:
\begin{itemize}
    \item You should answer \answerYes{}, \answerNo{}, or \answerNA{}.
    \item \answerNA{} means either that the question is Not Applicable for that particular paper or the relevant information is Not Available.
    \item Please provide a short (1–2 sentence) justification right after your answer (even for NA). 
\end{itemize}

{\bf The checklist answers are an integral part of your paper submission.} They are visible to the reviewers, area chairs, senior area chairs, and ethics reviewers. You will be asked to also include it (after eventual revisions) with the final version of your paper, and its final version will be published with the paper.

The reviewers of your paper will be asked to use the checklist as one of the factors in their evaluation. While "\answerYes{}" is generally preferable to "\answerNo{}", it is perfectly acceptable to answer "\answerNo{}" provided a proper justification is given (e.g., "error bars are not reported because it would be too computationally expensive" or "we were unable to find the license for the dataset we used"). In general, answering "\answerNo{}" or "\answerNA{}" is not grounds for rejection. While the questions are phrased in a binary way, we acknowledge that the true answer is often more nuanced, so please just use your best judgment and write a justification to elaborate. All supporting evidence can appear either in the main paper or the supplemental material, provided in appendix. If you answer \answerYes{} to a question, in the justification please point to the section(s) where related material for the question can be found.

IMPORTANT, please:
\begin{itemize}
    \item {\bf Delete this instruction block, but keep the section heading ``NeurIPS Paper Checklist"},
    \item  {\bf Keep the checklist subsection headings, questions/answers and guidelines below.}
    \item {\bf Do not modify the questions and only use the provided macros for your answers}.
\end{itemize}


\begin{enumerate}

\item {\bf Claims}
    \item[] Question: Do the main claims made in the abstract and introduction accurately reflect the paper's contributions and scope?
    \item[] Answer: \answerYes{} 
    \item[] Justification: The abstract and the introduction reflect paper's contribution and scope: using three Large Language Models (R1-Distilled, Llama, Qwen) to generate computational cognitive models in four cognitive domains (Learning, Decision making, Planning and Memory), with LLM-generated models matching or surpassing the best handcrafted models reported in literature. These claims match results reported in the figures and results section of experiments 1-4. Additionally, we reference the control results in the abstract (performing miscellaneous analyses to assert robustness of our results), which are reported in the control experiment results section.

\item {\bf Limitations}
    \item[] Question: Does the paper discuss the limitations of the work performed by the authors?
    \item[] Answer: \answerYes{} 
    \item[] Justification: We  discuss limitations including our approach being applied to only four cognitive domains that are not reflective of the entire field of cognitive science, potential susceptibility of LLMs to prompt formulation, limited exploration of feedback options, etc. 

\item {\bf Theory assumptions and proofs}
    \item[] Question: For each theoretical result, does the paper provide the full set of assumptions and a complete (and correct) proof?
    \item[] Answer: \answerNA{} 
    \item[] Justification: Our work is not theoretical, therefore this does not apply.

    \item {\bf Experimental result reproducibility}
    \item[] Question: Does the paper fully disclose all the information needed to reproduce the main experimental results of the paper to the extent that it affects the main claims and/or conclusions of the paper (regardless of whether the code and data are provided or not)?
    \item[] Answer: \answerYes{} 
    \item[] Justification: We provide 1) the prompt structure that generalized across the domains we tested, as well as fully detailed prompts we used for each experiment 2) details on LLMs we used to run experiments and which parameters we set (e.g., temperature), 3) model fitting instructions (e.g., which optimizer we used and for how many random starting points), 4) examples of text version of different data sets. All the data and the code are available in online repositories. We will make them available before the conference. 
    

\item {\bf Open access to data and code}
    \item[] Question: Does the paper provide open access to the data and code, with sufficient instructions to faithfully reproduce the main experimental results, as described in supplemental material?
    \item[] Answer: \answerYes{} 
    \item[] Justification: All the data and the code are available in online repositories. We will make them available before the conference. 

\item {\bf Experimental setting/details}
    \item[] Question: Does the paper specify all the training and test details (e.g., data splits, hyperparameters, how they were chosen, type of optimizer, etc.) necessary to understand the results?
    \item[] Answer: \answerYes{} 
    \item[] Justification: We explain that for each domain we used three separate parts of the data set for constructing the LLM prompt, evaluating the LLM-generated models during sampling, and comparing the best LLM-generated models to the baselines. We will make the data and the code used for splitting train/evaluation/test sets available before the conference.

\item {\bf Experiment statistical significance}
    \item[] Question: Does the paper report error bars suitably and correctly defined or other appropriate information about the statistical significance of the experiments?
    \item[] Answer: \answerYes{} 
    \item[] Justification: We perform appropriate statistical tests for our results, and also explain the meaning of the error bars (e.g. whether it's across the runs or participants in the data set)

\item {\bf Experiments compute resources}
    \item[] Question: For each experiment, does the paper provide sufficient information on the computer resources (type of compute workers, memory, time of execution) needed to reproduce the experiments?
    \item[] Answer: \answerYes{} 
    \item[] Justification: We specify in the Methods section that for each of our experiments running an experiment took a maximum of eight hours on four Nvidia A100s with 40GB memory each.  
    
\item {\bf Code of ethics}
    \item[] Question: Does the research conducted in the paper conform, in every respect, with the NeurIPS Code of Ethics \url{https://neurips.cc/public/EthicsGuidelines}?
    \item[] Answer: \answerYes{} 
    \item[] Justification: The research conducted in the paper conforms with the NeurIPS Code of Ethics.

\item {\bf Broader impacts}
    \item[] Question: Does the paper discuss both potential positive societal impacts and negative societal impacts of the work performed?
    \item[] Answer: \answerNA{} 
    \item[] Justification: Our research is foundational and not tied to particular applications or deployments. 
    
\item {\bf Safeguards}
    \item[] Question: Does the paper describe safeguards that have been put in place for responsible release of data or models that have a high risk for misuse (e.g., pretrained language models, image generators, or scraped datasets)?
    \item[] Answer: \answerNA{} 
    \item[] Justification: Paper poses no such risks.

\item {\bf Licenses for existing assets}
    \item[] Question: Are the creators or original owners of assets (e.g., code, data, models), used in the paper, properly credited and are the license and terms of use explicitly mentioned and properly respected?
    \item[] Answer: \answerYes{} 
    \item[] Justification: We cited all original papers that produced the datasets we used.

\item {\bf New assets}
    \item[] Question: Are new assets introduced in the paper well documented and is the documentation provided alongside the assets?
    \item[] Answer: \answerNA{} 
    \item[] Justification: We did not collect new data for this paper, and all of our experiments are done in-context (i.e. without model training).

\item {\bf Crowdsourcing and research with human subjects}
    \item[] Question: For crowdsourcing experiments and research with human subjects, does the paper include the full text of instructions given to participants and screenshots, if applicable, as well as details about compensation (if any)? 
    \item[] Answer: \answerNA{} 
    \item[] Justification: We used data collected by other researchers for their original publications, and were not involved in the data collection protocol. We included the description of the tasks provided to the participants, in as much detail as is described in the original papers.

\item {\bf Institutional review board (IRB) approvals or equivalent for research with human subjects}
    \item[] Question: Does the paper describe potential risks incurred by study participants, whether such risks were disclosed to the subjects, and whether Institutional Review Board (IRB) approvals (or an equivalent approval/review based on the requirements of your country or institution) were obtained?
    \item[] Answer: \answerNA{} 
    \item[] Justification: We did not collect any new human data for our experiments. 

\item {\bf Declaration of LLM usage}
    \item[] Question: Does the paper describe the usage of LLMs if it is an important, original, or non-standard component of the core methods in this research? Note that if the LLM is used only for writing, editing, or formatting purposes and does not impact the core methodology, scientific rigorousness, or originality of the research, declaration is not required.
    \item[] Answer: \answerYes{} 
    \item[] Justification: Since LLMs are at the core of our experiments, we described in detail which LLMs we used (including what family, size, and how they were prompted).

\end{enumerate}

\end{document}